\documentclass[fleqn,10pt]{wlpeerj}
\usepackage{amssymb}
\usepackage{latexsym}

\usepackage{amsmath}
\usepackage{bm}
\usepackage{threeparttable}
\usepackage{algorithm}
\usepackage{algpseudocode}

\usepackage{hyperref}       
\usepackage{url}            

\usepackage{natbib}

\usepackage{booktabs}        
\usepackage{multirow}        
\usepackage{siunitx}         

\newcommand{\argmax}{\mathop{\rm arg~max}\limits}

\begin{document}

\title{AlphaViT: A flexible game-playing AI for multiple games and variable board sizes}
\author[1]{Kazuhisa Fujita}
\affil[1]{Komatsu University, 10-10 Doihara-Machi, Komatsu, Ishikawa, Japan 923-0921}
\corrauthor[1]{Kazuhisa Fujita}{kazu@spikingneuron.net}


\begin{abstract}
We present three game-playing agents incorporating Vision Transformers (ViT) into the AlphaZero framework: AlphaViT, AlphaViD (AlphaViT with a transformer decoder), and AlphaVDA (AlphaViD with learnable action embeddings). These agents can play multiple board games of varying sizes using a single neural network with shared weights, thus overcoming AlphaZero's limitation of fixed board sizes. AlphaViT employs only a transformer encoder, whereas AlphaViD and AlphaVDA incorporate both a transformer encoder and a decoder. In AlphaViD, the decoder processes outputs from the encoder, whereas AlphaVDA uses learnable embeddings as the decoder inputs. The additional decoder in AlphaViD and AlphaVDA provides flexibility to adapt to various action spaces and board sizes. Experimental results show that the proposed agents, trained on either individual games or on multiple games simultaneously, consistently outperform traditional algorithms, such as Minimax and Monte Carlo Tree Search. They approach the performance of AlphaZero despite relying on a single deep neural network (DNN) with shared weights. In particular, AlphaViT performs strongly across all evaluated games. Furthermore, fine-tuning the DNN with weights pre-trained on small board games accelerates convergence and improves performance, particularly in Gomoku. Interestingly, simultaneous training on multiple games yields performance comparable to, or even surpassing, that of single-game training. These results indicate the potential of transformer-based architectures for developing more flexible and robust game-playing AI agents that excel in multiple games and dynamic environments.
\end{abstract}

\flushbottom
\maketitle
\thispagestyle{empty}

\section{Introduction}
\label{sec:introduction}

Artificial intelligence (AI) has advanced remarkably in recent years, demonstrating its potential across a wide range of applications. One area where AI has excelled is mastering board games, often outperforming top human players. Notable achievements include AI agents that have defeated humans in games, such as Checkers, Chess \citep{Campbell:2002}, and Othello \citep{Buro:1997}. A significant turning point occurred in 2016 when AlphaGo \citep{Silver:2016}, an AI designed specifically for the game of Go, defeated the world's top players. Subsequently, AlphaZero \citep{Silver:2018} was introduced, demonstrating its capability to master various board games, including Chess, Shogi, and Go. These achievements further highlight AI's superhuman skill in this domain.

Despite these successes, many current game-playing AI agents suffer from a fundamental limitation: they are typically designed for only a single specific game and cannot play other games. Even within the same game, these agents cannot handle variations in board size. In contrast, humans can easily switch between different board sizes. For example, beginners in Go often start practicing on smaller boards (e.g., $9 \times 9$) before progressing to larger boards (e.g., $19 \times 19$). However, AI agents like AlphaZero, which are designed for a single specific game and fixed board size, require substantial reprogramming to accommodate such changes.

For AlphaZero, this limitation arises from its deep neural network (DNN) architecture, which requires a fixed input size. AlphaZero's DNN consists of residual blocks and multilayer perceptrons (MLPs) and is designed for a fixed input size. The output size of the residual blocks varies with changes in input size, creating inconsistencies with the expected MLP input size. Consequently, AlphaZero fails even with small changes in board size.

To address this limitation, we propose replacing residual blocks in the AlphaZero framework with Vision Transformer (ViT) \citep{Dosovitskiy:2021}. ViT is an image-classification DNN based on the transformer architecture. ViT divides an image into patches and infers its class from the patches. A key advantage of ViT is its flexibility in handling various image sizes. This flexibility enables the AlphaZero framework to adapt to various games and board sizes.

This paper presents game-playing agents capable of handling multiple games and variable board sizes using a single DNN. These agents, named AlphaViT, AlphaViD (AlphaViT with a transformer decoder), and AlphaVDA (AlphaViD with learnable action embeddings), are based on the AlphaZero framework. The agents predict the value of a game state and policy using a DNN, and choose moves via Monte Carlo Tree Search (MCTS) \citep{Browne:2012,Winands:2017}. Our computational experiments show that the proposed agents can be trained to play three games (Connect 4, Gomoku, and Othello) simultaneously using a single DNN with shared weights. Moreover, the proposed agents outperform traditional algorithms, such as Minimax and MCTS, across various games, while approaching the performance of AlphaZero, whether trained on a single game or multiple games simultaneously. Our goal is not to surpass the state-of-the-art strength of specialized single-game agents, but rather to demonstrate that a single transformer architecture can flexibly handle multiple games and board sizes.

Portions of this text were previously published as part of a preprint (https://arxiv.org/abs/2408.13871).

\section{Related work}

Game-playing AI agents have reached superhuman performance levels in traditional board games such as Checkers \citep{Schaeffer:1993}, Othello \citep{Buro:1997,Buro:2003}, and Chess \citep{Campbell:1999,Hsu:1999,Campbell:2002}. In 2016, AlphaGo \citep{Silver:2016}, a Go-playing AI, defeated the world's top Go players, marking the first superhuman-level performance in Go. AlphaGo relied on supervised learning from a large database of expert human moves and self-play data. Subsequently, AlphaGo Zero \citep{Silver:2017} defeated AlphaGo, relying only on self-play data. In 2018, \cite{Silver:2018} proposed AlphaZero, which has no restrictions on playable games. AlphaZero outperformed other superhuman-level AIs in Go, Shogi, and Chess. Interestingly, AlphaZero's capabilities extend beyond traditional two-player perfect information games, with research exploring its potential in more complex scenarios. For example, \cite{Hsueh:2018} showed AlphaZero's potential in nondeterministic games. Other extensions include handling continuous action spaces \citep{Moerland:2018} and supporting multiplayer games \citep{Petosa:2019}. However, AlphaZero cannot play multiple games simultaneously, nor can it handle game variants with the same rules but different board sizes using a single DNN with shared weights. This limitation is due to AlphaZero's DNN architecture for policy and value estimation. AlphaZero's DNN consists of residual blocks and MLPs. While the residual blocks can process input images of various sizes, the MLPs require fixed input sizes. Consequently, the DNN is not sufficiently flexible to accommodate variations in board size.

Researchers have improved the AlphaZero framework to overcome its limitations. \cite{Wu:2019} tackled some of these limitations by improving the efficiency of AlphaZero-like training in Go. Wu's model introduced techniques such as {\it playout cap randomization} and {\it policy target pruning}, which significantly accelerate self-play learning. Wu used global pooling layers to standardize varying board sizes to a fixed size, enabling the model to estimate values using MLPs effectively. Furthermore, Wu exclusively utilized convolutional layers for policy estimation, thereby eliminating the need for MLPs. This design enables seamless handling of varying board sizes while efficiently calculating the policy. This innovation represents a significant step toward creating more flexible AI agents capable of adapting to different board configurations. Similarly, \cite{Soemers:2023} explored transfer learning across various board games, employing fully convolutional networks with global pooling to enable effective transfer between games with different board sizes, shapes, and rules. Their approach has demonstrated the potential of convolutional architectures, enhanced with global pooling, to generalize across various game scenarios.

To address these limitations, we propose integrating the transformer architecture into the AlphaZero framework. The transformer architecture, initially developed for natural language processing \citep{Vaswani:2017}, has proven remarkably effective across domains, including image-processing tasks. Transformer-based models achieve exceptional performance in various image-related tasks, such as image classification \citep{Dosovitskiy:2021}, semantic segmentation \citep{Xie:2021}, video classification \citep{Li:2022}, and video captioning \citep{Zhao:2022}. ViT, introduced by Dosovitskiy et al. \citep{Dosovitskiy:2021}, is a remarkable example of a transformer-based model for image processing. ViT achieved state-of-the-art performance in image classification at the time of its introduction. A key feature of ViT is its independence from the input image size \citep{Dosovitskiy:2021}. Unlike convolutional neural networks, which require fixed-size inputs, ViT can process images of various sizes. It achieves this by dividing each image into fixed-size patches, each treated as a token in the transformer architecture. This flexibility makes ViT highly adaptable and efficient when handling different image sizes. By incorporating the ViT architecture into the AlphaZero framework, we aim to extend its capabilities to handle various games with different board sizes and enhance its flexibility.

Some researchers have explored the use of transformers in game-playing AI. For example, \cite{Czech:2024} proposed a variant of the AlphaZero framework, AlphaVile, using an original network architecture with modified lightweight transformer blocks and an original loss function. AlphaVile achieved better performance than AlphaZero in Chess. \cite{Ruoss:2024} reached grandmaster level in Chess with a transformer model trained solely on a dataset, without any self-play or game-tree search. \cite{Monroe:2024} proposed a transformer-based architecture for Chess-playing AI. Using a simple transformer architecture, they trained agents that reached a high level of play. These single-game successes suggest that a transformer architecture can replace convolutions; in this study, we take the next step and show that a transformer can simultaneously master multiple games and board sizes with a single set of weights.

\section{AlphaViT, AlphaViD, and AlphaVDA}
\label{sec:methods}

\begin{figure}[htbp]
  \begin{center}
      \includegraphics[width=0.6\linewidth]{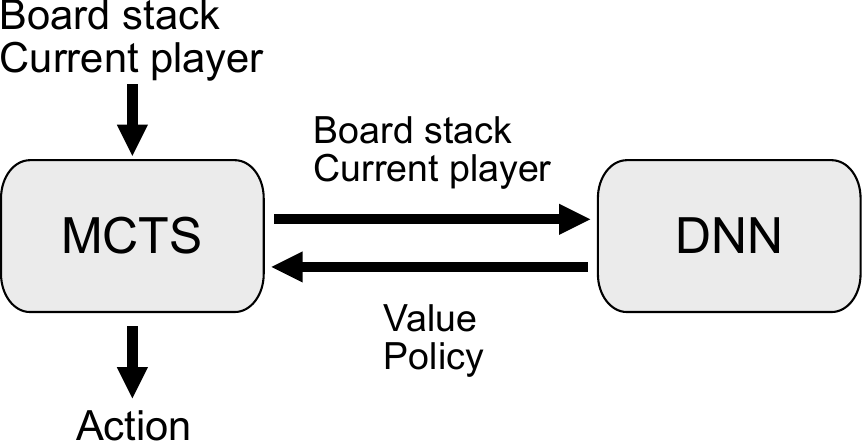}
      \caption{Overview of the decision process. The agents take a stack of board planes and the current player as input and determine the next move using Monte Carlo Tree Search (MCTS). MCTS explores the game tree using the value and policy probabilities provided by the DNN.}
      \label{fig:flow}                
  \end{center}
\end{figure}

AlphaZero's game-playing capability is limited to games with specific board sizes and rules used during training, as discussed in previous sections. This limitation arises from AlphaZero's DNN architecture, in which MLPs require fixed input sizes. To address this limitation, we propose AlphaViT, AlphaViD, and AlphaVDA as game-playing AI agents based on the AlphaZero framework but using ViT architecture. These agents use a combination of a DNN and MCTS (Figure \ref{fig:flow}). The DNN receives the board state and outputs a value estimate and move probabilities (policies). The MCTS searches a game tree using the estimated value and move probabilities. By incorporating ViT instead of residual blocks, AlphaViT, AlphaViD, and AlphaVDA can overcome the limitations of AlphaZero and play games that have variable board sizes and rule sets. While AlphaViT employs only a transformer encoder, AlphaViD and AlphaVDA employ both a transformer encoder and a decoder. Importantly, AlphaViT, AlphaViD, and AlphaVDA can play any game that AlphaZero can because they employ the same game-playing algorithm as AlphaZero.

The training method for AlphaViT, AlphaViD, and AlphaVDA is identical to that of AlphaZero, consisting of three stages: self-play, augmentation, and update. In the self-play stage, the agent generates training data by playing games against itself. The augmentation stage applies data augmentation techniques to the training data. Finally, during the update stage, the DNN weights are updated using the augmented training data. The details of the training method are described in Appendix \ref{sec:training}.

\subsection{Architectures of DNNs}

%

\paragraph{AlphaViT}

\begin{figure}[htbp]
  \begin{center}
      \includegraphics[width=0.7\linewidth]{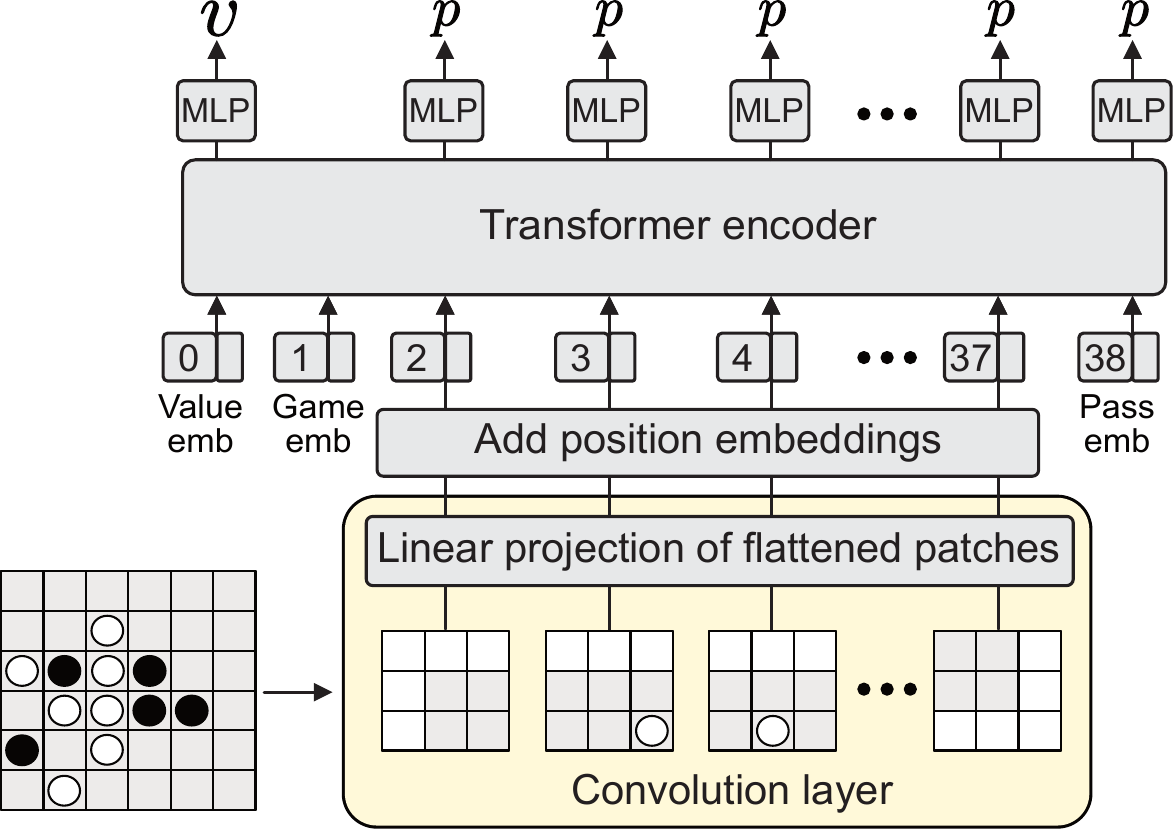}
      \caption{AlphaViT architecture. The input board is divided into patches by a convolutional layer, projected to embeddings, and combined with value, game, and pass tokens. After adding position embeddings, the sequence is processed by a transformer encoder. Outputs are used by MLP heads to predict the state value and move probabilities.}
      \label{fig:alphavit}                
  \end{center}
\end{figure}

An overview of AlphaViT's DNN architecture is shown in Figure \ref{fig:alphavit}. The DNN in AlphaViT is based on ViT, which has no input-size limitation and can classify images even if the input image size differs from the training image size. This flexibility enables AlphaViT to play games with different board sizes using the same network. In AlphaViT, the game boards are fed into ViT. Initially, these inputs are transformed into patch embeddings through a convolutional layer. Using a convolutional layer allows easy adjustment of the patch-division parameters, such as patch size, stride, and padding. The final output of the encoder is used to compute the value and the move probabilities. AlphaViT employs three special learnable embeddings (tokens) to enable flexibility for different games and board sizes:
\begin{itemize}
  \item \textbf{Value token} $\mathbf{x}_{\mathrm{value}}$: is consumed by an MLP value head to predict the state value $v(s)\in(-1,1)$, where $s$ is the game state, $1$ indicates a certain win, and $-1$ a certain loss.
  \item \textbf{Game token} $\mathbf{x}_{\mathrm{game}}$: encodes which game is currently being played. We use one-hot vectors to represent the game types (e.g., Connect 4, Gomoku, Othello).
  \item \textbf{Pass token} $\mathbf{x}_{\mathrm{pass}}$: represents the ``pass'' action that exists in Othello but not in Gomoku or Connect 4. By appending a dedicated learnable vector after all patch embeddings, we can predict the probability of the pass move when the rules require it.
\end{itemize}

Given a game state $s$, AlphaViT's DNN predicts the value $v(s)$ and the move probability vector $\bm p(s)$, which includes $p(a \mid s)$ for each action $a$. In a board game, $s$ represents the board state, and $a$ denotes a move.

The input to the DNN is an $H \times W \times (2T + 1)$ image stack ${\mathbf x} \in \mathbb{R}^{H \times W \times (2T + 1)}$, consisting of $2T + 1$ binary feature planes of size $H \times W$. Here, $H$ and $W$ are the dimensions of the board, and $T$ is the number of history planes. The first $T$ feature planes represent the occupancies of the first player's discs, where a feature value of 1 means that a disc occupies the corresponding cell, and 0 means it does not. The following $T$ feature planes represent the occupancies of the second player's discs. The final feature plane represents the color of the current player's disc, where 1 denotes the first player and -1 denotes the second player.

The convolutional layer divides the image stack ${\mathbf x}$ into $P \times P$ patches with stride $\mathrm{str}$ and padding $\mathrm{pad}$, and generates the patch embeddings $\mathbf{z}_\mathrm{patch}$. The patch size $P$ corresponds to the kernel size of the convolutional layer. The sequence of patch embeddings $\mathbf{z}_\mathrm{patch}$ is defined in Eq. \ref{eq:patch_embedding} as follows:
\begin{eqnarray}
  \label{eq:patch_embedding}
  \mathbf{z}_\mathrm{patch} = [{\mathbf x}^0_p {\mathbf E}; \ldots; {\mathbf x}^i_p {\mathbf E}; \ldots; {\mathbf x}^{N_p-1}_p {\mathbf E}], 
\end{eqnarray}
where $N_p$ is the number of patches, ${\mathbf x}^i_p \in \mathbb{R}^{P^2(2T + 1)}$ is the $i$th flattened 2D patch, and ${\mathbf E}$ is a trainable embedding tensor with the shape $(P^2(2T+1), D)$. Here, $N_p = \bigl ( \bigl\lfloor \frac{H + 2\,\mathrm{pad} - P}{\mathrm{str}} \bigr\rfloor + 1 \bigr ) \times \bigl ( \bigl\lfloor \frac{W + 2\,\mathrm{pad} - P}{\mathrm{str}} \bigr\rfloor + 1 \bigr )$, where $H$ and $W$ are the board height and width, $P$ is the patch size, $\mathrm{str}$ is the stride, and $\mathrm{pad}$ is the padding used in the convolutional layer. The kernel of the convolutional layer acts as the tensor ${\mathbf E}$, which maps each patch to a $D$-dimensional embedding space. Here, $D$ is the embedding size.

To retain position information, learnable 2D position embeddings ${\mathbf E}_\mathrm{pos}$ are added to the patch embeddings. These positional embeddings are scaled according to the board size and hyperparameters to match the patch embeddings $\mathbf{z}_\mathrm{patch}$. The resulting position-aware patch embeddings are given by Eq. \ref{eq:pos_embedding}:
\begin{eqnarray}
  \label{eq:pos_embedding}
  \mathbf{z}^\mathrm{pos}_\mathrm{patch} = \mathbf{z}_\mathrm{patch} + {\mathbf E}_\mathrm{pos}.
\end{eqnarray}

The output size of the transformer encoder is determined by the number of input embeddings. For Gomoku, where the action space is $HW$, AlphaViT requires $HW$ embeddings (i.e., $N_p = HW$). To achieve this, we set the patch size $P = 2k + 1$, where $k$ is a non-negative integer, stride $\mathrm{str} = 1$, and padding $\mathrm{pad} = \lfloor P /2 \rfloor$. Note that when $k > 0$, patches overlap, but the total number of patches remains $HW$. In contrast, for Othello, the action space is $HW+1$ to include the pass move. Using the same parameters as Gomoku ($P = 2k + 1$, $\mathrm{str} = 1$, and $\mathrm{pad} = \lfloor P /2 \rfloor$), AlphaViT requires one additional embedding for the pass move. To address this, we introduce a learnable pass embedding ${\mathbf x}_\mathrm{pass}$ by appending it after all patch embeddings (see Eq. \ref{eq:input_embedding}). To estimate the board value, we prepend a learnable value embedding ${\mathbf x}_\mathrm{value}$. Additionally, to enable AlphaViT to recognize different game types, we incorporate a static game embedding ${\mathbf x}_\mathrm{game}$, represented using one-hot encoding. These embeddings are appended to the embeddings $\mathbf{z}^\mathrm{pos}_\mathrm{patch}$. As a result, the input embeddings $\mathbf{z}_0$ to the transformer encoder are defined in Eq. \ref{eq:input_embedding} as follows:
\begin{eqnarray}
  \label{eq:input_embedding}
  \mathbf{z}_0 = [{\mathbf x}_\mathrm{value}; {\mathbf x}_\mathrm{game}; \mathbf{z}^\mathrm{pos}_\mathrm{patch}; {\mathbf x}_\mathrm{pass}].
\end{eqnarray}
Position embeddings are added before appending the pass, value, and game embeddings. This approach ensures that the position embeddings can be scaled independently, without being affected by embeddings that do not inherently contain positional information.

The sequence $\mathbf{z}_0$ is fed into the transformer encoder, which consists of $L$ transformer encoder layers. The output of the final encoder layer $\mathbf{z}_L$ has shape $(HW+3) \times D$. The first vector $\mathbf{z}^0_L$ derived from the value embedding is processed by the value head implemented as a multilayer perceptron (MLP) denoted as $\text{MLP}_v$. This head estimates the value $v$ using Eq. \ref{eq:value_head}:
\begin{equation}
  \label{eq:value_head}
  v = \tanh(\mathrm{MLP}_v(\mathrm{LN}(\mathbf{z}^0_L))),
\end{equation}
where LN represents layer normalization. The tanh activation constrains the value within the range $(-1, 1)$, representing the state value, where $1$ indicates a certain win and $-1$ a certain loss.

The vectors $\mathbf{z}_p = [\mathbf{z}^{2}_L, ..., \mathbf{z}^{HW + 2}_L]$ (index 0: value, 1: game, $2 \ldots HW+1$: patches, $HW+2$: pass), derived from the board patches and the pass embedding, are processed by the policy head, which is implemented as another MLP ($\mathrm{MLP}_p$). We flatten the $H \times W$ board positions and the pass move into one-dimensional indices $i \in \{0,1,\dots,HW\}$. For $i < HW$, the move corresponds to board coordinates $(m,n)$ where $m = i \mod W$ is the column index, and $n = \lfloor i / W \rfloor$ is the row index. The policy head applies a shared MLP to each token, and a Softmax over the resulting logits yields a probability vector $\bm p(s) \in \mathbb R^{HW+1}$ whose $i$-th entry is $p(a_i \mid s)$. The special index $i = HW$ represents the pass action. Thus, the policy head output is defined in Eq. \ref{eq:policy_head}:
\begin{equation}
  \label{eq:policy_head}
 p(a_i \mid s) = \mathrm{Softmax}(\mathrm{MLP}_{p}(\mathrm{LN}(\mathbf{z}_p)))_i,
\end{equation}
where $a_i$ is the $i$th action, and MLP is applied across all $HW + 1$ tokens. For Othello, the probability for the pass move is $p(a_{HW} \mid s)$. In Gomoku, this entry is masked out since the pass move is not allowed. In Connect 4, only the probabilities $\{p(a_i \mid s) \mid 0 \leq i < W\}$  (i.e., moves $(m=i, n=0)$) corresponding to valid column choices $i$ are used and renormalized. Note that $i$ is the column index in Connect 4. The agent drops the disc into column $i$, and the disc then falls to the lowest available row.

To decide on a move, AlphaViT employs MCTS with Upper Confidence Bound applied to Trees (UCT), using the value and move probabilities computed by the DNN. The MCTS algorithm used in AlphaViT is identical to that of AlphaZero, as detailed in Appendix \ref{sec:alphazero}. The hyperparameters for AlphaViT are listed in Appendix \ref{sec:parameters}.

\paragraph{AlphaViD and AlphaVDA}

\begin{figure}[htbp]
  \begin{center}
      \includegraphics[width=0.6\linewidth]{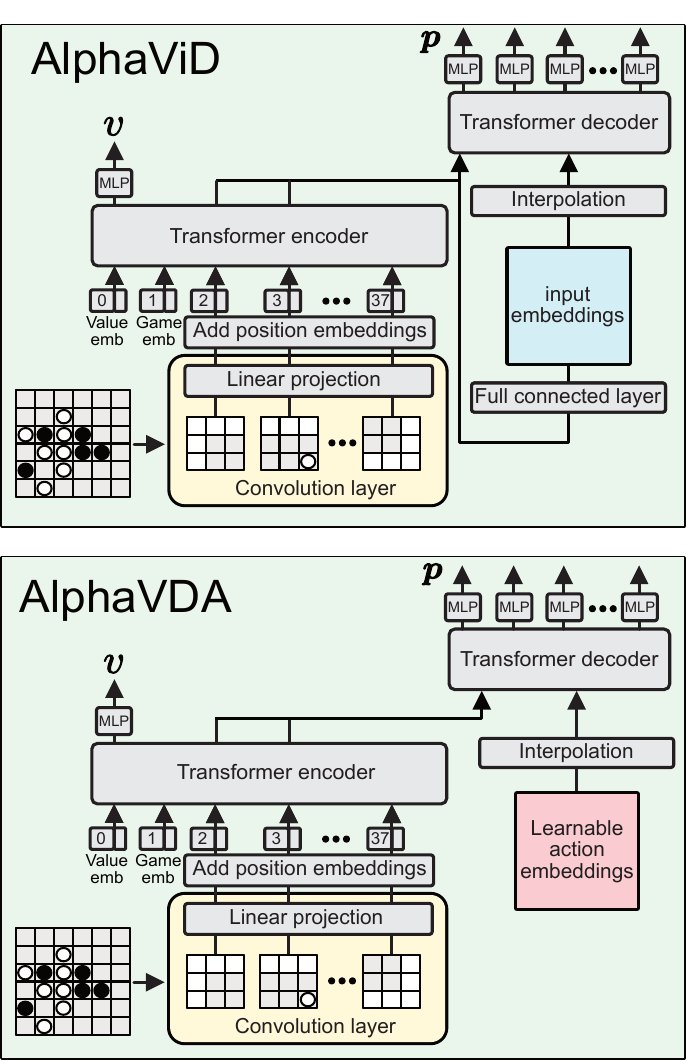}
      \caption{
        Architecture diagrams of AlphaViD (top) and AlphaVDA (bottom). The input board is divided into patches, embedded, and combined with value and game tokens, then processed by a transformer encoder. For policy prediction, encoder outputs are sent to a transformer decoder via interpolation. In AlphaViD, decoder inputs are generated from board-derived embeddings via a fully connected layer; in AlphaVDA, learnable action embeddings are used. MLP heads output the final value and move probabilities.}
      \label{fig:alphavid}                
  \end{center}
\end{figure}

AlphaViT's DNN has a significant drawback: the size of the policy vector is fixed by the transformer encoder's input size. To address this issue, AlphaViD and AlphaVDA incorporate both a transformer encoder and a decoder for calculating value and move probabilities, as shown in Figure \ref{fig:alphavid}. In AlphaViD, the decoder receives the input derived from the output of the encoder. In contrast, AlphaVDA employs learnable action embeddings as inputs to the decoder. The final output of the encoder is used to compute the value, while the final output of the decoder is used to compute the move probabilities. 

In AlphaViD and AlphaVDA, the input board is linearly embedded using a convolutional layer and fed into a transformer encoder, similar to AlphaViT. However, unlike AlphaViT, no pass embedding is included in the encoder input. The input embedding sequence is defined in Eq. \ref{eq:input_embedding_alphavid} as follows:
\begin{eqnarray}
  \label{eq:input_embedding_alphavid}
  \mathbf{z}_0 = [{\mathbf x}_\mathrm{value}; {\mathbf x}_\mathrm{game}; \mathbf{z}^\mathrm{pos}_\mathrm{patch}],
\end{eqnarray}
where the sequence consists of a value embedding, a game embedding, and patch embeddings. The embedding sequence in AlphaViD and AlphaVDA differs from that in AlphaViT. The estimated value is obtained from the value head that processes the output embedding corresponding to the value embedding from the last layer of the transformer encoder. In DNNs of AlphaViD and AlphaVDA, the encoder output is only used directly for value estimation, indicating that the number of patch embeddings $N_p$ does not need to match the action-space size.

The architecture of AlphaViD's DNN is shown at the top of Figure \ref{fig:alphavid}. The DNN estimates the move probability using the transformer decoder and MLP$_p$. The input embeddings for the transformer decoder are derived from the outputs of the transformer encoder corresponding to the patch embeddings, which are further processed through a fully connected layer. The initial embeddings for the decoder input are defined in Eq. \ref{eq:input_embedding_decoder} as follows:
\begin{eqnarray}
  \label{eq:input_embedding_decoder}
 {\mathbf y}'_{0} = \mathrm{MLP}([\mathbf{z}^2_L; \ldots;\mathbf{z}^{N_p + 1}_L]),\quad {\mathbf y}'_{0} \in \mathbb{R}^{N_p \times D_d},
\end{eqnarray}
where $D_d$ is the embedding size of the decoder. Since the embedding sequence size must match the input size of the transformer decoder, ${\mathbf y}'_{0}$ is interpolated to ${\mathbf y}_{0} \in \mathbb{R}^{N_a\times D_d}$, where $N_a$ is the action space size. In this study, this step uses bilinear interpolation in the function {\it torch.nn.functional.interpolate}. This interpolation provides flexibility to adjust the action space size depending on the game type and board size. If an additional move such as a ``pass'' (e.g., in Othello) is required, we simply set $N_a = HW + 1$ and resize ${\mathbf y}'_{0}$ accordingly via this interpolation. The additional embedding at the last position is assigned to the additional move (e.g., the pass). Therefore, a dedicated pass embedding is not required for the encoder's input. Similar to the original transformer, the transformer decoder receives ${\mathbf y}_{0}$ as the target sequence and $\mathbf{z}_{L}$ as the memory (encoder output). Finally, MLP$_p$ calculates the move probabilities from the output of the last layer of the transformer decoder ${\mathbf y}_L$. The probability of action $a_i$ is given by Eq. \ref{eq:alphavid_policy}:
\begin{equation}
  \label{eq:alphavid_policy}
 p(a_i \mid s) = \mathrm{Softmax}(\mathrm{MLP}_{p}(\mathrm{LN}({\mathbf y}_L)))_i.
\end{equation}

The architecture of AlphaVDA's DNN is shown at the bottom of Figure \ref{fig:alphavid}. The DNN is similar to AlphaViD's but uses learnable embeddings ${\mathbf y}'_{0}$ as the initial embeddings for the decoder input to the transformer decoder. While the length of these learnable embeddings is fixed, they are interpolated to the decoder input ${\mathbf y}_{0}$ to match the action space size, ensuring compatibility and adaptability across different game configurations.

To decide on a move, AlphaViD and AlphaVDA employ MCTS with UCT, using the value and move probabilities predicted by their DNNs. The MCTS algorithm implemented in these agents is identical to that of AlphaZero, as detailed in Appendix \ref{sec:alphazero}. The hyperparameters for AlphaViD and AlphaVDA are listed in Appendix \ref{sec:parameters}.

\section{Experimental setup}

\subsection{Games}

This study evaluates our agents on six game variants (three different games, each played on two board sizes): 
Connect 4 ($7\times 6$; ``Connect 4'', and $5\times 4$; ``Connect 4 5x4''), Gomoku ($9\times 9$; ``Gomoku'', and $6\times 6$; ``Gomoku 6x6''), and Othello ($8\times 8$; ``Othello'', and $6\times 6$; ``Othello 6x6''). These games are two-player, deterministic, zero-sum games with perfect information. Connect 4, published by Milton Bradley, is a connection game played on a $7\times 6$ board. The players take turns dropping discs onto the board. A player wins by forming a straight line of four discs horizontally, vertically, or diagonally. Connect 4 5x4 is a variant of Connect 4 with a $5\times 4$ board. Gomoku is a connection game in which players place stones on a board to form a straight line of five stones, either horizontally, vertically, or diagonally. This study uses a $9 \times 9$ board for Gomoku and a $6 \times 6$ board for Gomoku 6x6. Othello (Reversi) is a two-player strategy game played on an $8 \times 8$ board. In Othello, the disc is white on one side and black on the other. Players take turns placing a disc with their assigned color facing up. During a game, discs of the opponent's color are flipped to the current player's color if they are in a straight line and bounded by the disc just placed and another disc of the current player's color. Othello 6x6 is played on a $6 \times 6$ board in this study. We deliberately start with Connect 4, Gomoku, and Othello because (i) each has natural board-size variants (5x4 -- 9x9), (ii) they are commonly used for benchmarking, and (iii) training can be completed on commodity GPUs.

\subsection{Opponents}

This study evaluates the performance of AlphaViT, AlphaViD, and AlphaVDA using five different AI methods: AlphaZero, two variants of MCTS labeled MCTS100 and MCTS400, Minimax, and Random. AlphaZero is trained using the method described in Appendix \ref{sec:alphazero}. The MCTS methods (MCTS100 and MCTS400) were run with 100 and 400 simulations, respectively. 

Previous studies indicate that vanilla MCTS with random roll-outs scales poorly, remaining weaker than shallow $\alpha$--$\beta$ search even with increased the number of simulations. For instance, in Connect-4, a UCT agent employing $10000$ random simulations achieved only a $19.8\,\%$ win rate against a depth-8 $\alpha$--$\beta$ opponent \citep{Scheiermann:2023}. In contrast, incorporating domain-specific heuristics such as ``decisive-move'' pruning substantially improves MCTS performance, often yielding strength increases of one to two orders of magnitude \citep{Teytaud:2010,Taylor:2024}. Our measurements (Appendix \ref{sec:mcts_experiment}) corroborate these findings for MCTS, showing rapid Elo improvement up to approximately $400$ simulations followed by diminishing returns. Therefore, we adopt $100$ and $400$ simulations as the MCTS baselines in our experiments, providing clearly separated strength levels while maintaining reasonable computational requirements.

The details of MCTSs are provided in Appendix \ref{sec:mcts}. In these MCTS methods, the child nodes are expanded at the fifth visit to a node. Minimax selects a move using the minimax algorithm based on the evaluation table described in Appendix \ref{sec:minimax}. Random selects moves uniformly at random from valid moves.

\subsection{Software}

AlphaViT, AlphaViD, AlphaVDA, the opponents, and the board games were implemented in Python, using NumPy for linear algebra operations and PyTorch for DNNs. The source code is available on GitHub at \url{https://github.com/KazuhisaFujita/AlphaViT} for reproducibility and further extension of this work.

\subsection{Hardware}

All experiments were conducted on multiple custom-built PCs. The agents used a variety of consumer CPUs, including AMD Ryzen 9 9900X and Intel Core i7-14700K, Core i9-13900K, and Core i9-12900K processors. The experimental GPUs were NVIDIA GeForce RTX 4060 Ti 16 GB, RTX 3060 12 GB,  RTX 2070 Super, and RTX 2070. Each machine was equipped with 64 GB of RAM and two GPUs, and the experiments were run on Debian Linux. The model training and evaluation were distributed across multiple GPUs on one machine using data parallelization. No cloud-based or high-performance computing clusters were used, and all computations were performed using local custom-built hardware. For example, one of the machines used for training and evaluation was equipped with an Intel Core i9-13900K CPU, 64 GB of RAM, and two NVIDIA GeForce RTX 4060 Ti 16 GB GPUs.

\section{Results}

The results section presents the performance and characteristics of AlphaViT, AlphaViD, and AlphaVDA which were trained on different games (Connect 4, Gomoku, and Othello) with two board sizes (large and small). The main architectural difference between these agents lies in the number of encoder layers, which directly affects their learning capacity. Table~\ref{tab:num_parms} shows the number of parameters for each agent configuration. Each agent was tested with different numbers of encoder layers, denoted by `L' followed by a number (e.g., L1, L4, L5, L8). The number of parameters ranges from 11.2 to 19.9 million, increasing with encoder depth. For comparison, the AlphaZero agent, which serves as the baseline agent, has 7.1 million parameters.

The primary architectural variable is the number of transformer encoder layers. Throughout this paper, we designate four layers for AlphaViT and one layer for AlphaViD and AlphaVDA ($\approx 11$ million parameters) as the baseline configuration, which we also refer to as the shallower configuration in contrast to the deeper variants described later. By contrast, deeper encoders use exactly four additional layers (e.g., L8 for AlphaViT, L5 for AlphaViD/AlphaVDA, $\approx 20$ million parameters). Here, ``deeper'' is used only in this \emph{relative} sense and does not imply a universal threshold on the number of layers.

Each AI agent was trained on specific games with different board sizes. Table~\ref{tab:AlphaV_agents} categorizes the agents based on the games on which they were trained and the board sizes used during training. The first group includes agents trained on a single game with a large board, denoted as LB. The second group consists of agents trained on a single game with a small board, denoted as SB. The third group comprises agents simultaneously trained on multiple games, including Connect 4, Gomoku, and Othello, with large boards, denoted as Multi. The agents in the third group were trained on the three games and can play these three games using a single DNN. In other words, they do not specialize in a specific game. This diversity of training settings allows us to evaluate the agents' adaptability and generalization capabilities across different game domains.

\begin{table*}[htbp]
  \centering
  \caption{Encoder Layer Variations and Parameter Sizes in AI Agents}
  \begin{tabular}{|c|c|c|}
  \hline
      AI agent        & Number of encoder layers & Number of parameters  \\ \hline
      AlphaViT L4     &                     4 & 11.2M             \\ \hline  
      AlphaViD L1     &                     1 & 11.5M             \\ \hline              
      AlphaVDA L1     &                     1 & 11.3M             \\ \hline
      AlphaViT L8     &                     8 & 19.6M             \\ \hline  
      AlphaViD L5     &                     5 & 19.9M             \\ \hline              
      AlphaVDA L5     &                     5 & 19.8M             \\ \hline
      AlphaZero       &                     - & 7.1M              \\ \hline
  \end{tabular}
  \label{tab:num_parms}
\end{table*}

\begin{table*}
  \centering
  \caption{Board Size and Game Variations in AI Agent Training}
  \begin{tabular}{|c|c|c|}
  \hline
  AI agents                                         & Game                      & Board size \\ \hline
  AlphaViT LB, AlphaViD LB, AlphaVDA LB             & one specific game         & Large      \\ \hline
  AlphaViT SB, AlphaViD SB, AlphaVDA SB             & one specific game         & Small      \\ \hline    
  AlphaViT Multi, AlphaViD Multi, AlphaVDA Multi    & Connect 4, Gomoku, Othello & Large      \\ \hline  
  \end{tabular}
  \label{tab:AlphaV_agents}
\end{table*}

\subsection{Baseline Elo ratings}
\label{sec:results_Elo}

\paragraph{Objective and Setup.}

Tables \ref{tab:Connect4_elo}, \ref{tab:gomoku_elo}, and \ref{tab:othello_elo} present Elo ratings of various AI agents across different games and board sizes. Elo ratings provide a standard measure of relative performance in two-player games and enable systematic comparison across agents. We evaluated multiple variants of our proposed models, including agents trained on a single game with either large or small boards, and multiple games with large boards. Comparisons were conducted with other AI agents, including AlphaZero, MCTS with different numbers of simulations, Minimax, and a Random agent. Our proposed agents and AlphaZero underwent 1000 training iterations; each iteration consisted of self-play, augmentation, and an update step detailed in Appendix \ref{sec:training}. The Elo ratings of all agents were initialized to 1500 and were calculated through 50 round-robin tournaments between the agents. Each Elo rating is accompanied by a 95\% confidence interval (95\% CI); the calculation details are provided in Appendix \ref{sec:elo}.

Note: Throughout this paper, we use the term \textit{strong} to indicate benchmark-relative performance: an agent is considered strong if its Elo rating closely approaches or matches that of our AlphaZero (see Tables \ref{tab:Connect4_elo}, \ref{tab:gomoku_elo}, and \ref{tab:othello_elo}). Thus, \textit{strong} does not refer to an absolute numerical threshold but instead to AlphaZero-level performance.

\paragraph{Highlights across games.}

Single-task AlphaViT L4 approaches AlphaZero's performance across all evaluated games, remaining within approximately 270 Elo points of AlphaZero. Single-task AlphaViD L1 and AlphaVDA L1 achieve AlphaZero-level performance only on smaller boards (Connect 4 5x4 and Gomoku 6x6), yet they experience substantial drops in Elo ratings on larger boards. Multitask AlphaViT L4 (AlphaViT L4 Multi) demonstrates competitive performance on large boards and closely approaches AlphaZero (within 200 Elo points). Conversely, multitask AlphaViD L1 and AlphaVDA L1 variants exhibit significantly lower performance. Note that the multitask models were not trained on small board configurations and therefore underperform on this board size.

\paragraph{Connect 4 (Table \ref{tab:Connect4_elo}).}

For standard Connect 4, AlphaViT L4 Multi most closely approaches AlphaZero ($\Delta$ Elo $\approx -160$). AlphaViT L4 LB is moderately weaker ($\Delta$ Elo $\approx -260$), whereas AlphaViD L1 LB and AlphaVDA L1 LB lag behind significantly ($\Delta$ Elo $<-300$). On the smaller 5x4 board, all single-task variants (AlphaViT L4 SB, AlphaViD L1 SB, AlphaVDA L1 SB) are competitive with AlphaZero.

\paragraph{Gomoku (Table \ref{tab:gomoku_elo}).}

For the large board, only AlphaViTs (L4 LB and Multi) attain performance comparable to AlphaZero, with overlapping confidence intervals. AlphaViD and AlphaVDA variants (L1 LB and Multi) substantially lag behind ($\Delta$ Elo $< -300$). On the smaller board, SB variants from all architectures effectively approach AlphaZero-level performance.

\paragraph{Othello (shown in Table \ref{tab:othello_elo}).}

On the large board, AlphaViT L4 LB numerically surpasses AlphaZero ($\Delta$ Elo $\approx +20$), though overlapping confidence intervals prevent definitive conclusions. AlphaViD L1 LB and AlphaViT L4 Multi achieve performance near that of AlphaZero, with $\Delta$ Elo $\approx -100$. AlphaVDA L1 LB lags behind significantly ($\Delta$ Elo $<-300$). On the smaller board, AlphaViT L4 SB closely approaches AlphaZero ($\Delta$ Elo $\approx -70$). AlphaViD L1 SB lags behind ($\Delta$ Elo $\approx -230$), whereas AlphaVDA L1 SB also falls short ($\Delta$ Elo $\approx -180$), remaining weaker than AlphaZero. Interestingly, AlphaViT L4 LB, despite being trained only on the large board, achieves performance comparable to AlphaViD L1 SB and AlphaVDA L1 SB on the smaller board. This suggests that knowledge may transfer from larger to smaller boards.

\paragraph{Conclusion.}

In summary, AlphaViT L4 and AlphaViT L4 Multi consistently achieve or closely approach AlphaZero-level strength across games, even without game-specific fine-tuning. In contrast, AlphaViD L1 and AlphaVDA L1 achieve AlphaZero-level performance only on smaller boards. The strong results for multitask agents (AlphaViT L4 Multi) suggest that multitask training will not significantly hinder performance. The Elo ratings presented in Table \ref{tab:Connect4_elo}, \ref{tab:gomoku_elo}, and \ref{tab:othello_elo} serve as a baseline for the subsequent experiments described in the following sections.

\begin{table*}[htbp]
    \centering
    \caption{Elo ratings of AI agents for \textbf{Connect 4} variants}
    \setlength{\tabcolsep}{6pt}
    \renewcommand{\arraystretch}{1.1}
    \begin{tabular}{|l|r r r|r r r|}
    \hline
        \multirow{2}{*}{Agent} &
        \multicolumn{3}{c|}{Connect 4} &
        \multicolumn{3}{c|}{Connect 4 5x4} \\
        \cline{2-7}
                            & Elo       & $\Delta$ Elo& 95\% CI            & Elo       & $\Delta$ Elo& 95\% CI           \\ \hline\hline
        AlphaZero      & $ {\bf2114}$& -            &[$2073$,$2159$]   & ${\bf 1769}$ &         -  & [$1747$,$1797$]     \\ \hline
        AlphaViT L4 LB      & $ 1846     $    &$ -267.4$ &[$1799$,$1902$]   & $1517$       & $   -252$  & [$1480$,$1554$]    \\ \hline
        AlphaViD L1 LB      & $ 1739     $    &$ -374.7$ &[$1696$,$1783$]   & $1462$       & $ -306.9$  & [$1426$,$1495$]    \\ \hline
        AlphaVDA L1 LB      & $ 1746     $    &$ -367.3$ &[$1707$,$1788$]   & $1507$       & $ -262.1$  & [$1472$,$1539$]    \\ \hline
        AlphaViT L4 SB      & $ 1202     $    &$ -911.6$ &[$1152$,$1252$]   & ${\bf 1751}$ & $ -18.07$  & [$1722$,$1777$]    \\ \hline
        AlphaViD L1 SB      & $ 1177     $    &$ -936.4$ &[$1135$,$1214$]   & ${\bf 1764}$ & $ -4.512$  & [$1741$,$1793$]    \\ \hline  
        AlphaVDA L1 SB      & $ 1251     $    &$ -862.5$ &[$1201$,$1301$]   & ${\bf 1789}$ & $  19.98$  & [$1763$,$1818$]    \\ \hline
        AlphaViT L4 Multi   & $ 1950     $    &$ -164.0$ &[$1917$,$1988$]   & $1402$       & $ -366.4$  & [$1364$,$1437$]    \\ \hline  
        AlphaViD L1 Multi   & $ 1745     $    &$ -369.2$ &[$1706$,$1785$]   & $1317$       & $ -451.4$  & [$1271$,$1361$]    \\ \hline
        AlphaVDA L1 Multi   & $ 1669     $    &$ -444.7$ &[$1627$,$1710$]   & $1305$       & $ -463.3$  & [$1266$,$1343$]    \\ \hline
        MCTS400             & $ 1516     $    &$ -836.1$ &[$1467$,$1560$]   & $1563$       & $ -205.1$  & [$1530$,$1599$]    \\ \hline  
        MCTS100             & $ 1278     $    &$ -598.2$ &[$1230$,$1316$]   & $1502$       & $ -266.9$  & [$1468$,$1533$]    \\ \hline  
        Minimax             & $ 1012     $    &$  -1102$ &[$970.6$,$1053$]  & $1316$       & $   -453$  & [$1277$,$1346$]     \\ \hline      
        Random              & $755.4     $    &$  -1358$ &[$722.5$,$788.9$] & $1038$       & $ -730.4$  & [$992.4$,$1082$]    \\ \hline
    \end{tabular}
    \caption*{Elo ratings of AI agents for Connect 4 variants. Bolded ratings indicate agents whose Elo rating is within 100 points of AlphaZero. Elo ratings were calculated from 50 round-robin tournaments beginning with an initial rating of 1500, including differences relative to AlphaZero ($\Delta$ Elo) and corresponding 95\% confidence intervals.}
    \label{tab:Connect4_elo}
\end{table*}

\begin{table*}[htbp]
    \centering
    \caption{Elo ratings of AI agents for \textbf{Gomoku} variants}
    \setlength{\tabcolsep}{6pt}
    \renewcommand{\arraystretch}{1.1}
    \begin{tabular}{|l|r r r|r r r|}
    \hline
        \multirow{2}{*}{Agent} &
        \multicolumn{3}{c|}{Gomoku} &
        \multicolumn{3}{c|}{Gomoku 6x6} \\
        \cline{2-7}
                            & Elo        & $\Delta$ Elo& 95\% CI           & Elo         & $\Delta$ Elo& 95\% CI           \\ \hline\hline
        AlphaZero      & ${\bf 2038}$ &         -  & [$1982$,$2091$]   & ${\bf 1807}$ & -          &[$1788$,$1830$]  \\ \hline
        AlphaViT L4 LB      & ${\bf 1966}$ & $  -72.4$  & [$1924$,$2012$]   & $ 1645$      &$  -162$    &[$1611$,$1678$]    \\ \hline 
        AlphaViD L1 LB      & $  1698$    & $   -340$  & [$1659$,$1744$]   & $ 1509$       &$-298.3$    &[$1476$,$1546$]    \\ \hline
        AlphaVDA L1 LB      & $  1567$    & $ -471.5$  & [$1521$,$1614$]   & $ 1261$       &$-546.3$    &[$1219$,$1297$]    \\ \hline  
        AlphaViT L4 SB      & $  1553$    & $ -485.7$  & [$1500$,$1605$]   & ${\bf 1764}$ &$-42.78$    &[$1745$,$1786$]    \\ \hline
        AlphaViD L1 SB      & $  1519$    & $ -519.6$  & [$1466$,$1565$]   & ${\bf 1779}$ &$-27.43$    &[$1760$,$1798$]    \\ \hline 
        AlphaVDA L1 SB      & $ 977.2$    & $  -1061$  & [$943.8$,$1011$]  & ${\bf 1771}$ &$-35.96$    &[$1751$,$1792$]    \\ \hline
        AlphaViT L4 Multi   & ${\bf 2024}$& $ -13.93$  & [$1985$,$2069$]   & $ 1656$      &$-150.8$    &[$1626$,$1688$]    \\ \hline  
        AlphaViD L1 Multi   & $  1530$    & $ -508.7$  & [$1484$,$1576$]   & $ 1378$      &$-428.8$    &[$1339$,$1416$]    \\ \hline
        AlphaVDA L1 Multi   & $  1570$    & $ -467.8$  & [$1517$,$1622$]   & $ 1170$      &$-636.9$    &[$1130$,$1209$]    \\ \hline
        MCTS400             & $  1161$    & $ -877.4$  & [$1116$,$1201$]   & $ 1680$      &$-126.3$    &[$1651$,$1709$]    \\ \hline  
        MCTS100             & $  1229$    & $ -809.5$  & [$1180$,$1274$]   & $ 1380$      &$-426.3$    &[$1343$,$1418$]    \\ \hline     
        Minimax             & $  1472$    & $ -566.4$  & [$1425$,$1518$]   & $ 1348$      &$-458.6$    &[$1309$,$1384$]     \\ \hline      
        Random              & $ 696.8$    & $  -1341$  & [$673.9$,$719.2$] & $851.9$      &$-954.9$    &[$820.8$,$884.9$]   \\ \hline
    \end{tabular}
    \caption*{Elo ratings of AI agents for Gomoku variants. Bolded ratings indicate agents whose Elo rating is within 100 points of AlphaZero. Elo ratings were calculated from 50 round-robin tournaments beginning with an initial rating of 1500, including differences relative to AlphaZero ($\Delta$ Elo) and corresponding 95\% confidence intervals.}
    \label{tab:gomoku_elo}
\end{table*}

\begin{table*}[htbp]
    \centering
    \caption{Elo ratings of AI agents for \textbf{Othello} variants}
    \setlength{\tabcolsep}{6pt}
    \renewcommand{\arraystretch}{1.1}
    \begin{tabular}{|l|r r r|r r r|}
    \hline
        \multirow{2}{*}{Agent} &
        \multicolumn{3}{c|}{Othello    } &
        \multicolumn{3}{c|}{Othello 6x6} \\
        \cline{2-7}
                            & Elo       & $\Delta$ Elo& 95\% CI           & Elo       & $\Delta$ Elo& 95\% CI           \\ \hline\hline
        AlphaZero      & ${\bf 1996}$ &         -  & [$1953$,$2048$]   & ${\bf 2034}$ & -          &[$1986$,$2081$]   \\ \hline
        AlphaViT L4 LB      & ${\bf 2017}$ & $ 20.55$   & [$1973$,$2065$]   & $ 1819$      &$-215.5$    &[$1776$,$1863$]   \\ \hline        
        AlphaViD L1 LB      & ${\bf 1896}$ & $-99.95$   & [$1854$,$1941$]   & $ 1585$       &$-449.5$    &[$1539$,$1631$]   \\ \hline
        AlphaVDA L1 LB      & $  1669$    & $-327.2$   & [$1620$,$1718$]   & $ 1368$       &$-666.4$    &[$1322$,$1415$]   \\ \hline 
        AlphaViT L4 SB      & $  1482$    & $-513.9$   & [$1435$,$1528$]   & ${\bf 1963}$ &$-71.16$    &[$1919$,$2013$]   \\ \hline
        AlphaViD L1 SB      & $  1184$    & $-812.4$   & [$1132$,$1228$]   & $ 1803$      &$-231.6$    &[$1757$,$1854$]   \\ \hline
        AlphaVDA L1 SB      & $  1111$     & $-885.3$   & [$1063$,$1159$]   & $ 1855$     &$-179.6$    &[$1799$,$1911$]   \\ \hline
        AlphaViT L4 Multi   & ${\bf 1910}$ & $ -86.4$   & [$1865$,$1958$]   & $ 1090$     &$-944.2$    &[$1046$,$1134$]   \\ \hline 
        AlphaViD L1 Multi   & $  1668$     & $-328.7$   & [$1626$,$1709$]   & $ 1308$     &$-726.4$    &[$1264$,$1354$]   \\ \hline
        AlphaVDA L1 Multi   & $  1578$     & $  -418$   & [$1529$,$1626$]   & $991.1$     &$ -1043$    &[$938.4$,$1034$]  \\ \hline
        MCTS400             & $  1373$     & $-623.6$   & [$1329$,$1412$]   & $ 1579$     &$  -455$    &[$1535$,$1623$]   \\ \hline  
        MCTS100             & $  1194$     & $-802.7$   & [$1148$,$1241$]   & $ 1367$     &$-667.6$    &[$1324$,$1412$]   \\ \hline 
        Minimax             & $  1140$     & $-855.9$   & [$1089$,$1190$]   & $ 1363$     &$-670.9$    &[$1311$,$1411$]    \\ \hline      
        Random              & $ 782.8$     & $ -1213$   & [$744.8$,$816.5$] & $874.2$     &$ -1160$    &[$824$,$919$]      \\ \hline
    \end{tabular}
    \caption*{Elo ratings of AI agents for Othello variants. Bolded ratings indicate agents whose Elo rating is within 100 points of AlphaZero. Elo ratings were calculated from 50 round-robin tournaments beginning with an initial rating of 1500, including differences relative to AlphaZero ($\Delta$ Elo) and corresponding 95\% confidence intervals.}
    \label{tab:othello_elo}
\end{table*}

\subsection{Variation of Elo rating over training iteration}
\label{sec:variation_elo}

\begin{figure}[htbp]
  \begin{center}
      \includegraphics[height=16cm]{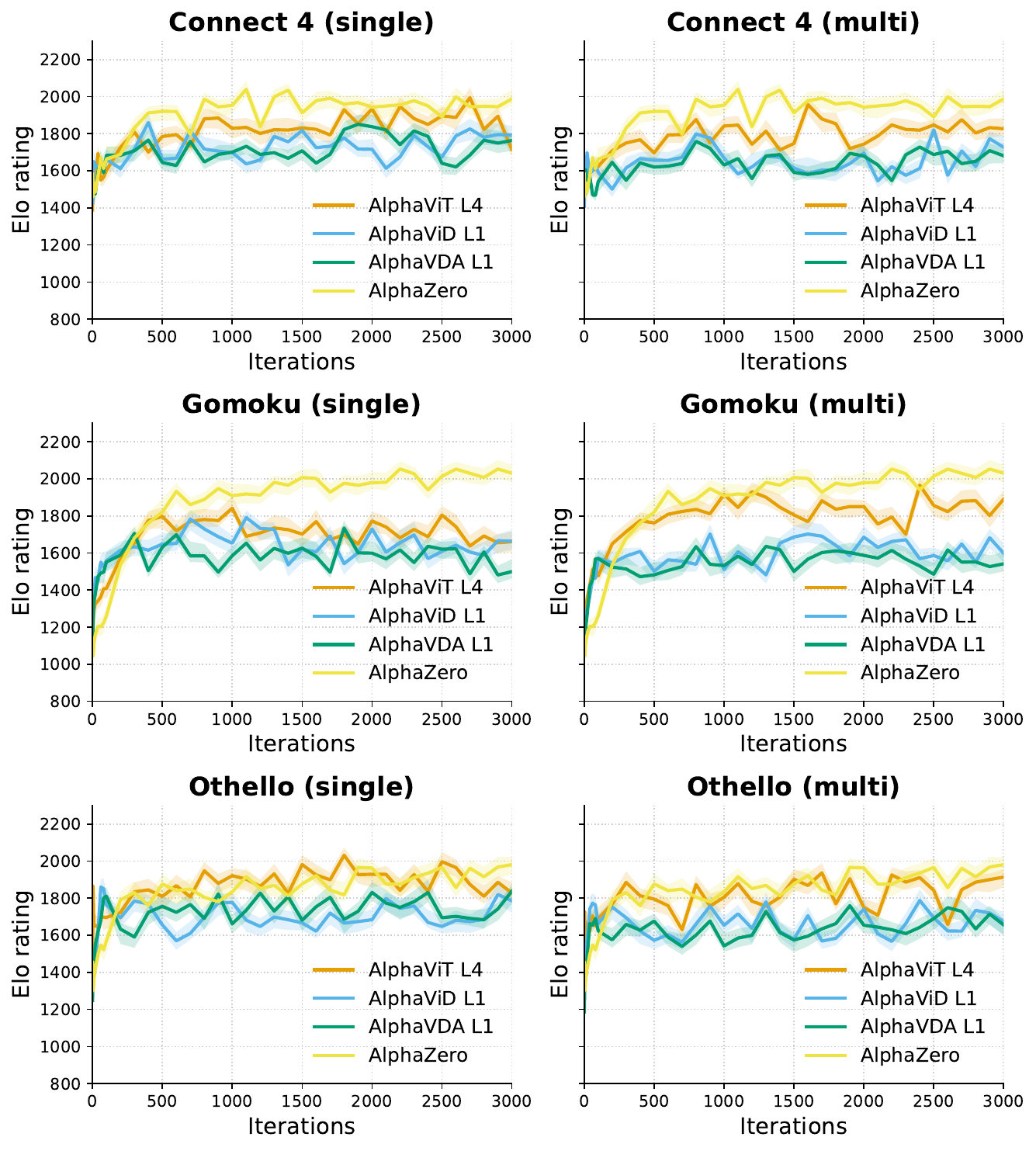}
      \caption{Elo rating progression over training iterations for AlphaViT, AlphaViD, and AlphaVDA in large board configurations (Connect 4, Gomoku, and Othello). The left column represents single-game-trained agents (single-task agents), whereas the right column indicates multi-game-trained agents (multitask agents). Solid lines represent Elo ratings calculated directly from aggregated game outcomes without employing bootstrapping, while shaded bands correspond to the 95\% confidence intervals around these Elo ratings.
      }
      \label{fig:elo_var_large}                
  \end{center}
\end{figure}

\begin{figure}[htbp]
  \begin{center}
      \includegraphics[height=16cm]{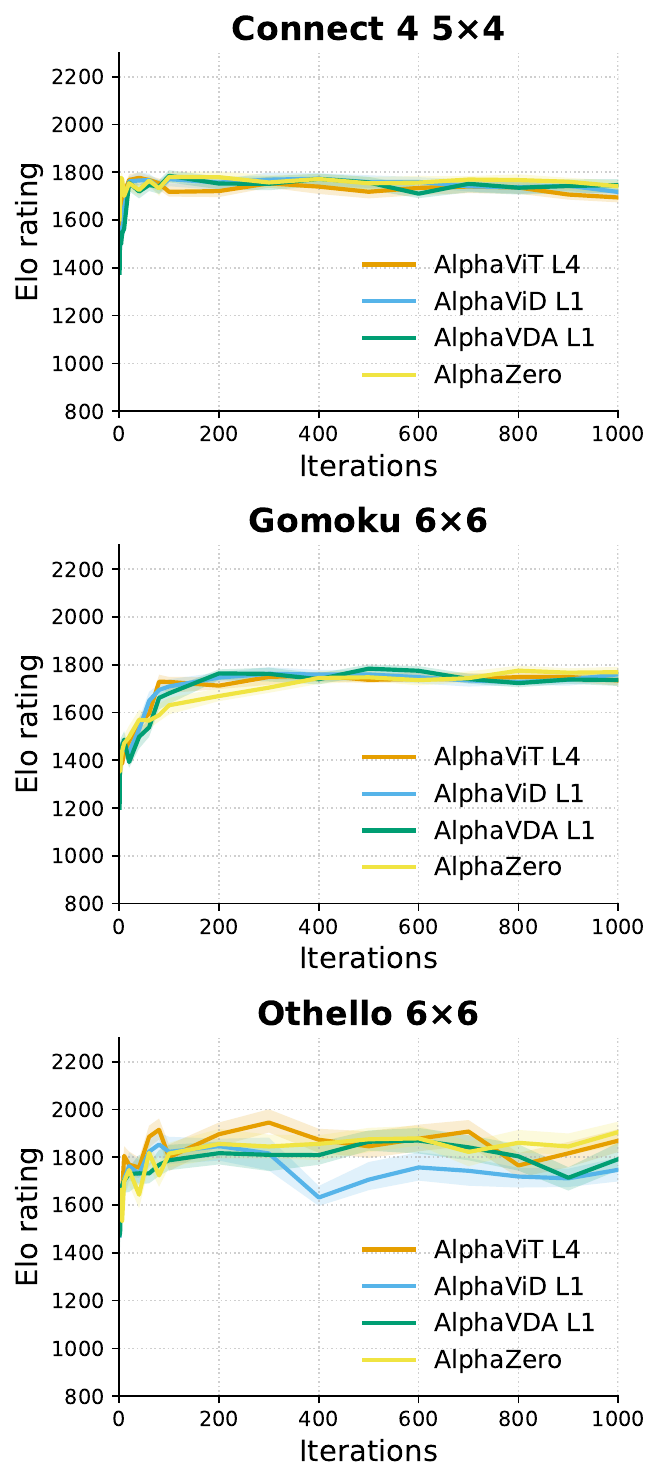}
      \caption{Elo rating progression over training iterations for AlphaViT, AlphaViD, and AlphaVDA in their small board configurations (Connect 4 5x4, Gomoku 6x6, and Othello 6x6). Solid lines represent Elo ratings calculated directly from aggregated game outcomes without employing bootstrapping, while shaded bands correspond to the 95\% confidence intervals around these Elo ratings.}
      \label{fig:elo_var_small}                
  \end{center}
\end{figure}

\paragraph{Setup.}
Figures \ref{fig:elo_var_large} and \ref{fig:elo_var_small} illustrate the progression of Elo ratings for AlphaViT, AlphaViD, AlphaVDA, and AlphaZero across multiple training iterations for large and small board configurations in three games: Connect 4, Gomoku, and Othello. Elo ratings were recorded at every iteration from 1 to 10, subsequently every 20 iterations up to iteration 100, and thereafter at intervals of 100 iterations until iteration 3000 for large board configurations and 1000 for small board configurations. The shaded areas around each Elo curve indicate the 95\% confidence intervals computed through bootstrapping, as detailed in Appendix \ref{sec:elo}. One iteration corresponds to a complete cycle of self-play data generation, data augmentation, and a single update of the neural network (see Appendix \ref{sec:training}). The opponents used for evaluation were identical to those described in previous experiments: AlphaViT L4 LB, AlphaViD L1 LB, AlphaVDA L1 LB, AlphaViT L4 SB, AlphaViD L1 SB, AlphaVDA L1 SB, AlphaViT L4 Multi, AlphaViD L1 Multi, AlphaVDA L1 Multi, AlphaZero, MCTS400, MCTS100, Minimax, and Random agents. The Elo ratings of these opponents were fixed as listed in Tables \ref{tab:Connect4_elo}, \ref{tab:gomoku_elo}, and \ref{tab:othello_elo}. Evaluated agents' Elo ratings were initialized to 1500 and then calculated through 40-game matches against each opponent, consisting of 20 games as the first player and 20 games as the second player.

\paragraph{Large board configurations.}
Across all tested large board games, AlphaViT, AlphaViD, and AlphaVDA variants demonstrate rapid Elo improvement during the initial training phase, achieving approximately 80--90 \% of their maximum Elo within the first 300--500 iterations. In Gomoku, the single-task agents peak near 1000 iterations and then exhibit a moderate decline. In Othello, the growth rate slows markedly after roughly 300 iterations, although the Elo ratings continue to increase slightly thereafter. Multitask-trained agents (AlphaViT Multi, AlphaViD Multi, AlphaVDA Multi) show learning speeds comparable to those of single-task models, confirming that multitask training does not adversely affect initial convergence. Notably, single-game-trained Gomoku models (AlphaViT L4 LB, AlphaViD L1 LB, AlphaVDA L1 LB) exhibit moderate Elo declines after reaching peak performance, indicating potential overfitting in later training stages. Conversely, multitask-trained Gomoku models (AlphaViT L4 Multi, AlphaViD L1 Multi, AlphaVDA L1 Multi) remain stable without a pronounced decline, highlighting multitask learning as a potentially effective approach for mitigating overfitting.

\paragraph{Small board configurations.}
Agents trained on smaller board sizes exhibit notably faster convergence, attaining peak Elo ratings within the first $\leq 300$ iterations ($\approx100$ for Connect 4 5x4). After this rapid convergence, in Connect 4 5x4 and Gomoku 6x6, Elo ratings remain stable with minimal fluctuation, suggesting that small board training reliably and quickly yields robust and stable agents. However, in Othello 6x6, Elo ratings show a slight fluctuation within $\approx \pm 100$.

\paragraph{Insights and implications.}
These findings clearly demonstrate that AlphaViT L4, AlphaViD L1, and AlphaVDA L1 quickly achieve high-performance levels on small boards. However, achieving similar performance stability on larger board configurations remains more challenging. The observed plateau in Elo improvement on large boards implies that simply extending training iterations beyond a certain point (around 1000 iterations) yields limited additional benefit. Therefore, to achieve further performance improvements on larger boards, enhancing model complexity—such as employing deeper transformer encoders—rather than solely prolonging training duration may be necessary.

\clearpage

\subsection{Effect of transformer encoder depth on performance}
\label{sec:variation_deep}

\begin{figure}[htbp]
  \begin{center}
      \includegraphics[height=16cm]{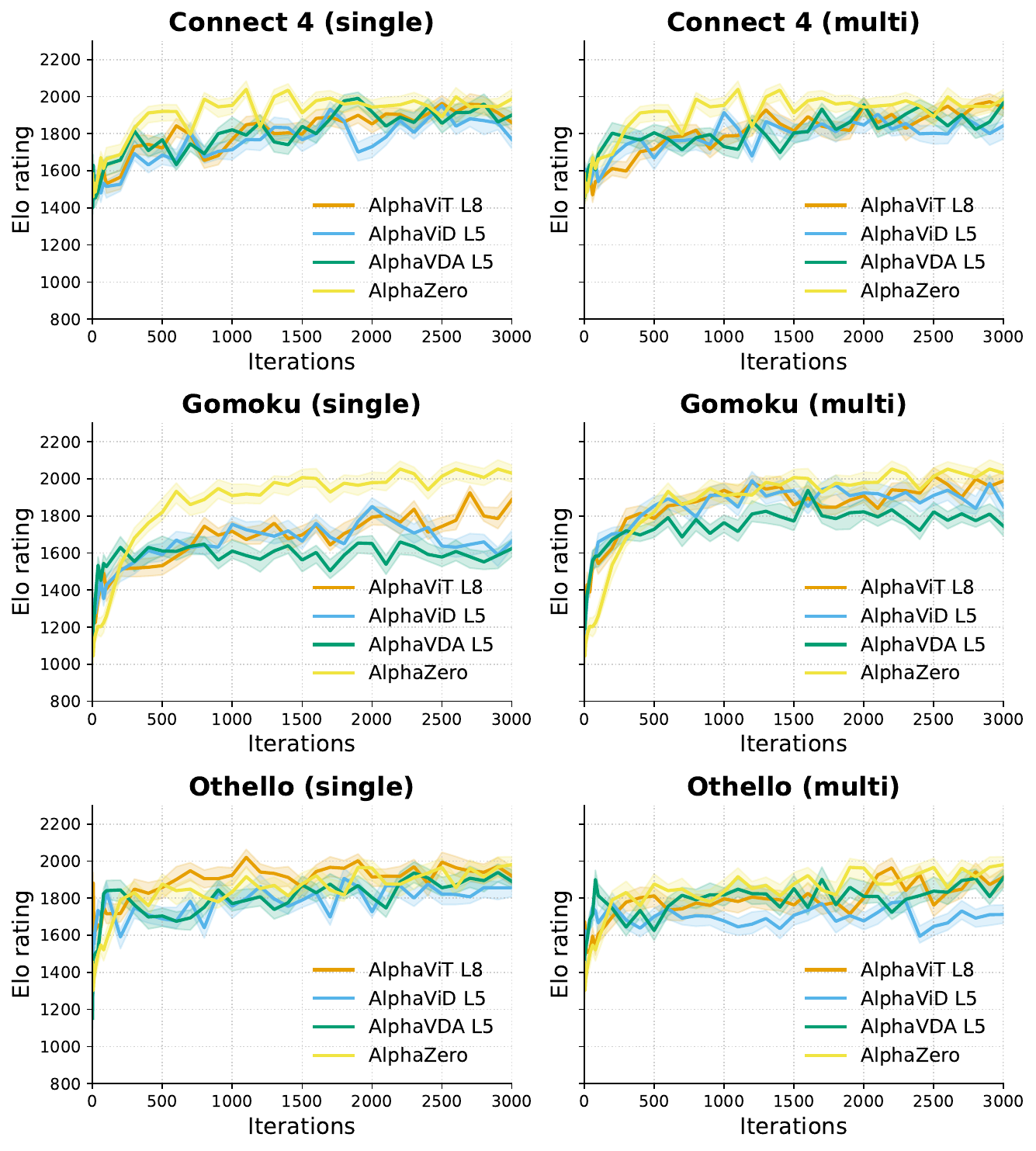}
      \caption{Elo trajectories for the deep configurations on standard boards. The left column represents single-game-trained agents, whereas the right column indicates multi-game-trained agents. Solid lines represent Elo ratings calculated directly from aggregated game outcomes without employing bootstrapping, while shaded bands correspond to the 95\% CIs around these Elo ratings.
      }
      \label{fig:elo_var_deep_large}                
  \end{center}
\end{figure}

\begin{figure}[htbp]
  \begin{center}
      \includegraphics[height=16cm]{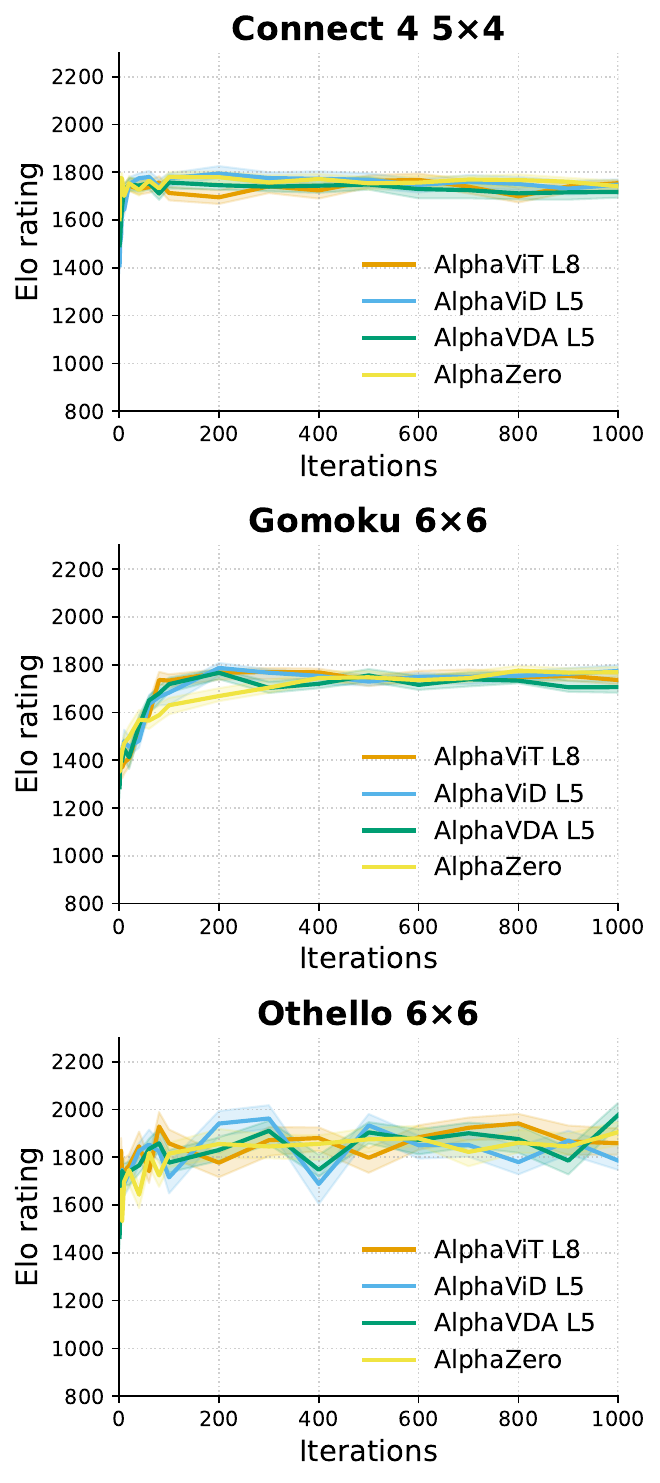}
      \caption{Elo trajectories for the deep configurations on small boards. Solid lines represent Elo ratings calculated directly from aggregated game outcomes without employing bootstrapping, while shaded bands correspond to the 95\% CIs around these Elo ratings.}
      \label{fig:elo_var_deep_small}                
  \end{center}
\end{figure}

In this subsection, we evaluated the performance of the agents with deeper encoders: AlphaViT L8, AlphaViD L5, and AlphaVDA L5. Figures \ref{fig:elo_var_deep_large} and \ref{fig:elo_var_deep_small} show the variations in Elo ratings for these agents over the iterations on large and small boards, respectively. All experiments followed the evaluation protocol described in the Setup paragraph of Section \ref{sec:variation_elo}.

For large board configurations, as illustrated in Figure \ref{fig:elo_var_deep_large}, the Elo ratings of agents with deeper encoders gradually stabilize between roughly 1000 and 2000 iterations. This trend indicates a more protracted improvement phase compared to their shallower counterparts (AlphaViT L4, AlphaViD L1, and AlphaVDA L1). Conversely, in small board configurations, depicted in Figure \ref{fig:elo_var_deep_small}, Elo ratings, as well as their 95\% CIs, converge more rapidly, similar to those of the shallow encoder agents. This suggests that in less complex game environments, the additional depth of encoders offers little substantial benefit.

Table \ref{tab:elos_2000-3000} lists mean Elo ratings averaged over iterations 2100--3000 for large board configurations. This highlights the performance gains achieved through the increased encoder depth, with AlphaViD and AlphaVDA demonstrating the most notable improvements. For example, AlphaViD L5 Multi and AlphaVDA L5 Multi achieve gains of $+286$ and $+231$ Elo points, respectively, in Gomoku, indicating significant performance enhancement.

\begin{figure}[htbp]
  \begin{center}
      \includegraphics[width=1\linewidth]{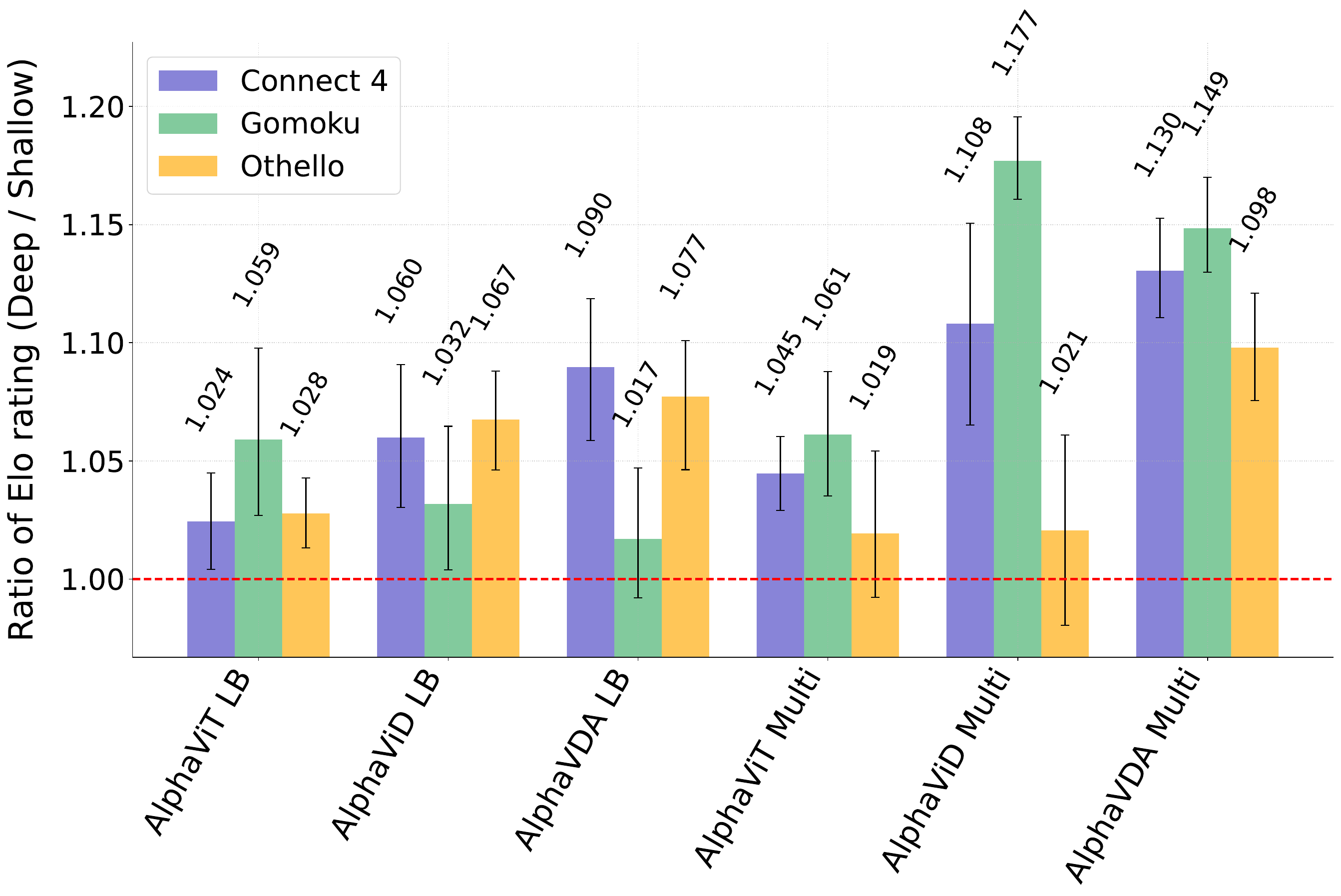}
      \caption{The ratio of Elo ratings between deeper and baseline DNNs for AlphaViT, AlphaViD, and AlphaVDA across Connect 4, Gomoku, and Othello. Ratios greater than 1.0 indicate superior performance of agents with deeper DNNs compared to those with baseline DNNs. The error bars represent 95\% CIs calculated through bootstrapping (see Appendix \ref{sec:elo}). }
      \label{fig:elo_ratio}                
  \end{center}
\end{figure}

The ratio of Elo ratings between the deeper and baseline DNNs, illustrated in Figure \ref{fig:elo_ratio}, further substantiates these findings. Across all evaluated games, deeper architectures generally exhibit superior performance, with AlphaViD Multi and AlphaVDA Multi achieving the highest ratios. Specifically, in Gomoku, these agents achieve Elo ratios of 1.177 and 1.149, respectively. They also outperform the baseline models in Connect 4, with Elo ratios of 1.108 and 1.130, respectively. While single-game-trained AlphaViD and AlphaVDA show marked gains in Connect 4 and Othello, their performance improvements in Gomoku are modest. In contrast, both single- and multi-game-trained AlphaViT variants exhibit relatively weaker improvements than AlphaViD and AlphaVDA. In conclusion, these results collectively demonstrate that increasing the depth of the transformer encoder layers positively influences agent performance across various game types and board configurations. This effect is particularly pronounced for AlphaViD and AlphaVDA, especially in larger and more complex game settings.

\begin{table*}[htbp]
  \centering
  \caption{Mean Elo Ratings from 2100 to 3000 iterations}
  \begin{tabular}{|c|c|c|c|}
  \hline
      Game              & Connect 4 ($\Delta$ Elo)     & Gomoku ($\Delta$ Elo)      & Othello ($\Delta$ Elo)   \\ \hline
      AlphaZero         & {\bf 1955}(-)    & {\bf 2019} (-)    & {\bf 1925} (-)    \\ \hline   
      AlphaViT L4 LB    & {\bf 1870} (-)   & 1703 (-)          & {\bf 1890} (-)    \\ \hline
      AlphaViT L8 LB    & {\bf 1915} (+45) & 1804 (+101)       & {\bf 1942} (+52)  \\ \hline
      AlphaViD L1 LB    & 1747 (-)         & 1630 (-)          & 1726 (-)    \\ \hline 
      AlphaViD L5 LB    & 1851 (+104)      & 1682 (+52)        & {\bf 1843} (+117) \\ \hline               
      AlphaVDA L1 LB    & 1738 (-)         & 1569 (-)          & 1749 (-)    \\ \hline
      AlphaVDA L5 LB    & {\bf 1894} (+156)& 1596 (+27)        & {\bf 1884} (+135) \\ \hline
      AlphaViT L4 Multi & 1827 (-)         & 1835 (-)          & {\bf 1848} (-)    \\ \hline  
      AlphaViT L8 Multi & {\bf 1909} (+82) & {\bf 1947} (+112) & {\bf 1883} (+35)  \\ \hline
      AlphaViD L1 Multi & 1658 (-)         & 1616 (-)          & 1669 (-)    \\ \hline
      AlphaViD L5 Multi & 1838 (+180)      & 1902 (+286)       & 1704 (+35)  \\ \hline                 
      AlphaVDA L1 Multi & 1666 (-)         & 1556 (-)          & 1670 (-)    \\ \hline
      AlphaVDA L5 Multi & {\bf 1884} (+218)& 1787 (+231)       & {\bf 1833} (+163) \\ \hline
  \end{tabular}
  \caption*{Mean Elo ratings from 2100 to 3000 iterations for each agent. $\Delta$ Elo indicates the difference from a shallower model. Bolded ratings indicate agents whose Elo rating is within 100 points of AlphaZero.}
  \label{tab:elos_2000-3000}
\end{table*}

\clearpage

\subsection{Effect of fine-tuning from small board games}
\label{sec:fine-tuning}

\begin{figure}[htbp]
  \begin{center}
      \includegraphics[width=1\linewidth]{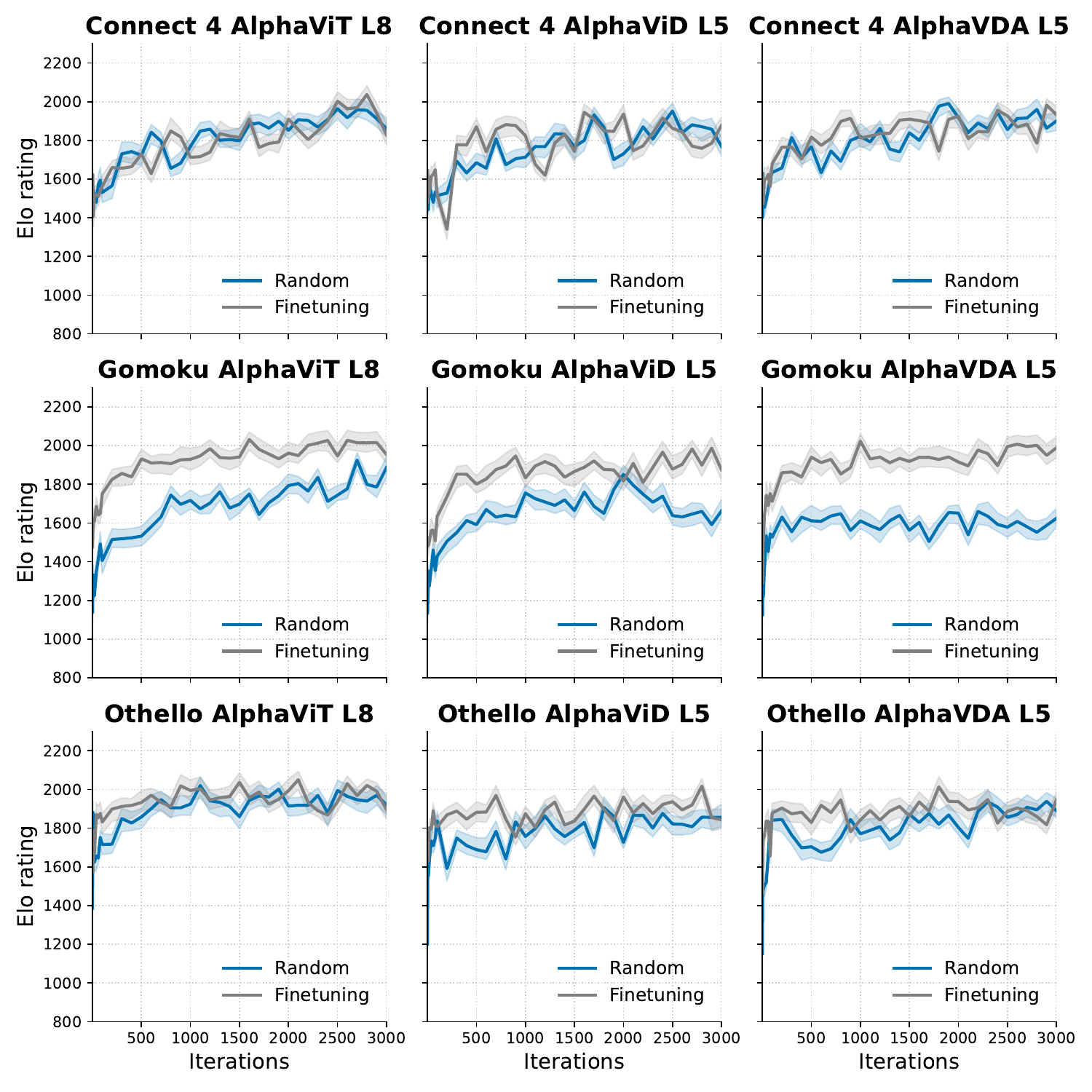}
      \caption{Elo rating progression over training iterations of the agents (AlphaViT L8, AlphaViD L5, and AlphaVDA L5) with fine-tuned and randomly initialized DNNs for three board games (Connect 4, Gomoku, and Othello). The three columns show the results for AlphaViT L8, AlphaViD L5, and AlphaVDA L5, respectively. The three rows show the results for Connect 4, Gomoku, and Othello. The shaded bands represent the 95\% confidence intervals around the Elo ratings, calculated through bootstrapping (see Appendix \ref{sec:elo}).}
      \label{fig:elo_var_fine}                
  \end{center}
\end{figure}

Figure \ref{fig:elo_var_fine} illustrates the Elo ratings of AlphaViT L8, AlphaViD L5, and AlphaVDA L5 with fine-tuned and non-fine-tuned (randomly initialized) DNNs across three games: Connect 4, Gomoku, and Othello. The fine-tuned DNNs were initialized using weights from the single-task DNNs that had been trained for 200 iterations on the small board configuration. For example, the fine-tuned DNN for Connect 4 was initialized with weights from the DNN trained on Connect 4 5x4. In contrast, agents with non-fine-tuned DNNs were initialized with random weights. All experiments followed the evaluation protocol described in the Setup paragraph of Section \ref{sec:variation_elo}. For agents with non-fine-tuned (randomly initialized) DNNs, we reused the Elo ratings previously reported in Section \ref{sec:variation_deep}.

The results demonstrate that agents with fine-tuned DNNs consistently achieve higher Elo ratings than those with randomly initialized DNNs for Gomoku and Othello. The performance difference is especially pronounced during the early iterations; in this phase, fine-tuned DNNs improve rapidly and then stabilize for Gomoku and Othello. For Gomoku, the advantage of fine-tuning is clear, as the agents with fine-tuned DNNs consistently outperform those with non-fine-tuned DNNs throughout the training process. The plateau appears in AlphaViD L5 and AlphaVDA L5 for Gomoku between approximately 1000 and 2000 iterations. However, the performance improves after 2000 iterations. For Connect 4, fine-tuning yields little or no improvement in Elo.

\section{Conclusion}

In this paper, we present AlphaViT, AlphaViD, and AlphaVDA, which are novel game-playing AI agents designed to overcome the limitations of AlphaZero using Vision Transformers (ViT). Unlike AlphaZero, which is restricted to fixed board sizes, our proposed agents demonstrate adaptability, handle different board sizes effectively, and exhibit flexibility across games. Furthermore, these agents can simultaneously train on and play multiple games, such as Connect 4, Gomoku, and Othello, within a single shared neural network. The performance of these multitask agents surpasses that of traditional game AI algorithms and, in some cases, approaches AlphaZero's performance.

The results of this study show that AlphaViT, AlphaViD, and AlphaVDA outperform traditional methods such as Minimax and MCTS across all tested scenarios. Although AlphaZero remains the top performer in some cases, particularly for games with larger boards, the proposed agents exhibit competitive performance. AlphaViT L8 matches AlphaZero in Connect 4 and Othello. In Othello, the deeper versions (L5) of AlphaViD and AlphaVDA narrow the gap with AlphaZero but do not yet surpass it. Multigame-trained variants perform on par with or better than single-game-trained variants with deeper DNNs in Connect 4 and Gomoku, while remaining slightly behind in Othello.

AlphaViT, AlphaViD, and AlphaVDA show strong adaptability across different games and board sizes. The agents with DNNs trained on a single game often achieve performance comparable to traditional game algorithms, such as Minimax and MCTS, even when playing on board sizes on which they are not trained. Moreover, multi-game-trained agents frequently perform on par with or surpass their single-game-trained counterparts. The agents with fine-tuned DNNs trained on small board games achieve performance better than that of agents with non-fine-tuned DNNs in Gomoku and Othello, but show little or no improvement in Connect 4. In the case of Gomoku, pre-trained weights from small board games significantly accelerate convergence and enhance the final performance. This suggests effective knowledge transfer between different board sizes, mirroring human learning processes, in which skills acquired in simpler variants (e.g., 9x9 Go) can be applied to more complex versions (e.g., 19x19 Go). Such adaptability suggests that the proposed agents may have significant potential for advancing the development of multitask AI. 

Comparing the three proposed agents reveals that AlphaViT L4 outperforms both AlphaViD L1 and AlphaVDA L1, despite having a similar number of parameters. This difference in performance may be attributed to the smaller number of encoder layers used in AlphaViD and AlphaVDA. This observation is further supported by the fact that AlphaViD L5 and AlphaVDA L5 exhibit performance comparable to AlphaViT L4. The simpler architecture of AlphaViT, consisting solely of encoder layers, may lead to more efficient performance in certain games, even when it has fewer parameters than its variants. However, this simplicity constrains AlphaViT's flexibility, as its output size is fixed to the number of input embeddings, limiting its applicability to games beyond classic board games. In contrast, including a decoder layer in AlphaViD and AlphaVDA allows for dynamic adjustment of the policy vector size, providing greater adaptability to games with varying action spaces. This architectural flexibility makes AlphaViD and AlphaVDA versatile candidates for handling more complex games or environments with continuous action spaces.

\section{Discussion}

Our results also relate to those of \cite{Soemers:2021,Soemers:2023}, who highlighted the importance of transferring trained policies and value functions across games with varying board sizes and action spaces. Similarly, our agents, equipped with DNNs that utilize weights either fine-tuned from a small board game or trained simultaneously on multiple games, demonstrate enhanced gameplay skills. This suggests that knowledge obtained from small board games and other games is used efficiently. The ability to train on multiple games concurrently may also help these agents avoid overfitting, further enhancing their generalization capabilities.

As shown in Tables \ref{tab:Connect4_elo}, \ref{tab:gomoku_elo}, and \ref{tab:othello_elo}, AlphaZero achieves the top performance with the smallest number of parameters in some cases, particularly for games with larger boards. This likely stems from the strong inductive bias inherent in its convolutional and residual network architecture, which is well-suited for single games and enables it to easily extract local and spatial features \citep{He:2016,Battaglia:2018}. Remarkably, this architecture also allows some flexibility in board size when game rules remain unchanged. For example, Wu's KataGo preserves the convolutional backbone while introducing global pooling layers—calculating channel-wise mean, scaled mean, and maximum—to both the trunk and heads, enabling a unified network to play Go from 9x9 through 19x19 without increasing the number of parameters \citep{Wu:2019}.

However, the strong inductive biases of CNNs may limit adaptability when game rules or mechanics significantly change. \cite{Soemers:2021} systematically demonstrated that fully convolutional networks can effectively transfer learning across different board sizes or minor variants of the same game but struggle significantly when transferring between games with substantially different rules or mechanics, resulting in poor or negative transfer performance even after fine-tuning. In contrast, transformer-based agents sacrifice some of this inductive bias in favor of greater flexibility \citep{Dosovitskiy:2021,dAscoli:2022}. This enables them to handle multiple games and variable board sizes at the cost of an increase in parameter count. In this study, when we employed deep transformer encoders and model parameters of AlphaViT, AlphaViD, and AlphaVDA were much larger than those of AlphaZero's DNN, their performance approached, and in Othello even surpassed, that of AlphaZero. This trade-off highlights a key design consideration: models with strong inductive bias can achieve higher efficiency in specialized domains, whereas more general architectures require additional capacity to compensate for their weaker assumptions \citep{dAscoli:2022}.

Although our proposed agents with transformer-based DNNs are slightly weaker than AlphaZero, we hypothesize they have the potential for further improvement through enhanced training techniques. For example, properly scheduled learning-rate decay during training iterations \citep{Silver:2016,Silver:2017,Silver:2018} and warm-start method \citep{Wang:2020,Wang:2021} will improve performance. Increasing the number of self-plays per iteration may also be effective.

All experiments were carried out on affordable, consumer-grade GPUs (RTX 4060 Ti 16 GB and RTX 3060 12 GB) rather than on datacenter accelerators. Due to the significant memory requirements for training transformer architectures, even moderately sized boards, such as Gomoku 9x9 and Othello, approached the memory limits of these GPUs. To address this issue, we enabled automatic mixed-precision training with {\it torch.cuda.amp.autocast()}, so that most tensor operations run in FP16/BF16, while numerically sensitive layers and master weights remain in FP32. Additionally, we employed data-parallel training across multiple GPUs within a single custom-built PC to further reduce per-device memory usage. These practical choices emphasize the reproducibility of AlphaViT but also restrict experimentation with deeper transformer encoders and larger boards, such as 19x19 Go. Nevertheless, the proposed architecture inherently supports variable input sizes, and our fine-tuning experiments (Section \ref{sec:fine-tuning}) indicate that weights pre-trained on small board games effectively generalize to larger board configurations. Consequently, when more powerful hardware is available, it will be feasible to first train agents with larger transformer encoders on small boards and then efficiently fine-tune, rather than training from scratch, to extend the proposed agents to full-sized boards and more complex games.

In future work, we plan to extend these architectures to a broader range of games, including those with more complex rules and stochastic elements. In addition, we aim to incorporate the flexibility of ViT-based DNNs into other deep reinforcement learning frameworks, such as deep Q-networks, to create AI agents capable of playing a wider variety of games, including video games, with enhanced adaptability.

\section*{Acknowledgments}

The author gratefully acknowledges the assistance of large language models in improving the grammar and style of this work.


\vspace{1cm}
\noindent\textbf{\textsf{\Large{Appendix}}}
\setcounter{section}{0}
\setcounter{figure}{0}
\setcounter{table}{0}

\renewcommand{\thesection}{\Alph{section}}
\renewcommand{\thefigure}{A\arabic{figure}}
\renewcommand{\thetable}{A\arabic{table}}

\section{AlphaZero}
\label{sec:alphazero}

AlphaZero integrates a deep neural network (DNN) with Monte Carlo Tree Search (MCTS), as illustrated in Figure \ref{fig:flow}. The DNN processes input representing the current board state and the current player, producing an estimated state value and a move-probability vector. MCTS then uses these outputs to select the optimal move. This same framework is adopted by AlphaViT, AlphaViD, and AlphaVDA.

\subsection{Deep Neural Network in AlphaZero}

\begin{figure}[htbp]
  \centering
    \includegraphics[width=0.5\linewidth]{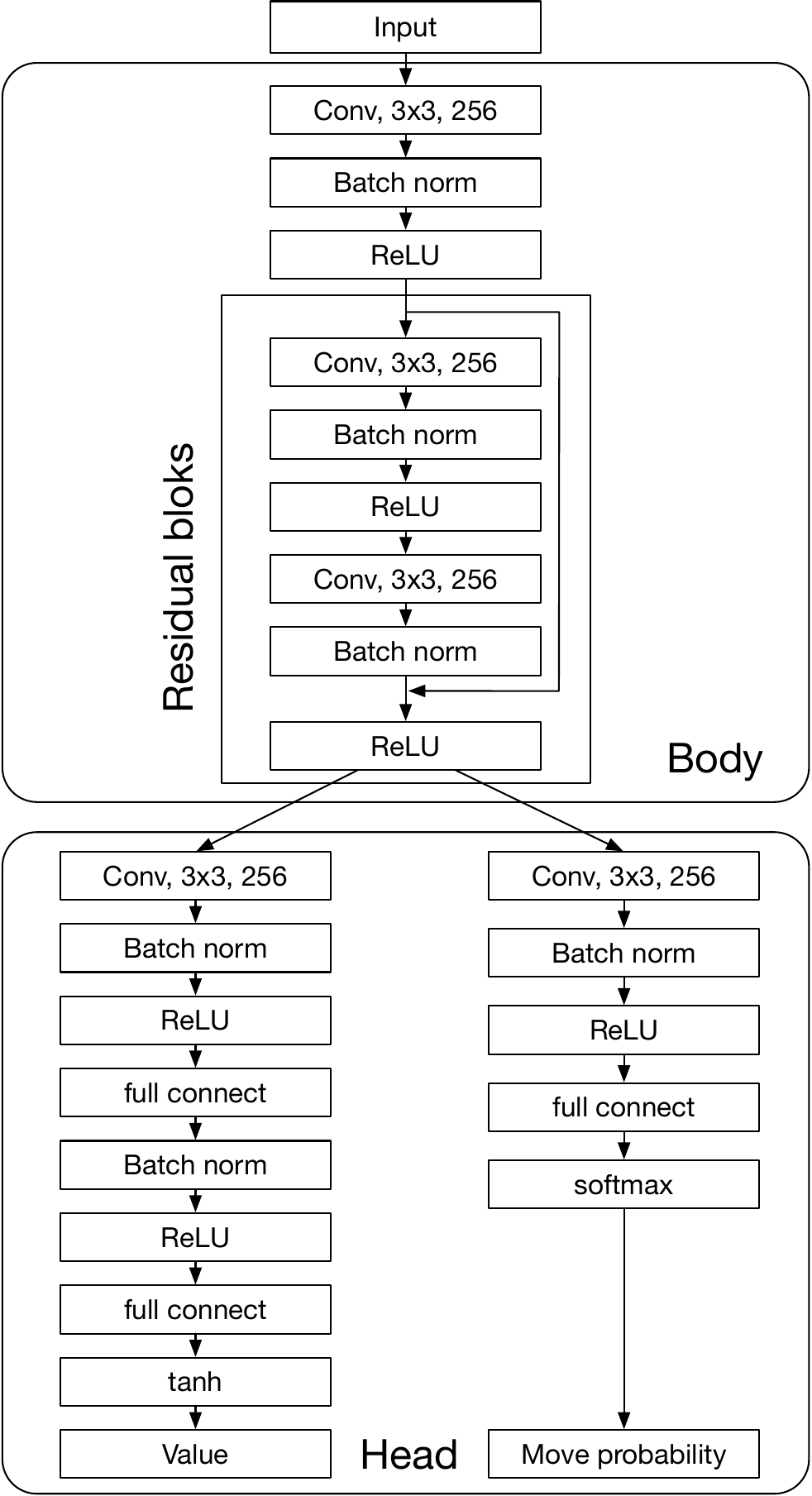}
  \caption{DNN of AlphaZero.}
  \label{fig:alphazero}
\end{figure}

AlphaZero's DNN predicts a value $v(s)$ and a move-probability vector $\bm{p}(s)$ with components $p(a \mid s)$ for each action $a$, given a state $s$. In the board game context, $s$ and $a$ represent the board state and the move, respectively. The DNN receives input representing the current board state and the current player's disc color. Figure \ref{fig:alphazero} illustrates the DNN architecture, which consists of a {\it Body} (residual blocks) and two {\it Heads} (value and policy heads). The value head outputs the estimated state value $v(s)$, while the policy head produces the move probabilities $\bm p(s)$.

The input to the DNN is an $H \times W \times (2T + 1)$ image stack that contains $2T + 1$ binary feature planes of size $H \times W$. Here, $H \times W$ refers to the board size, and $T$ is the number of histories (previous board states). The first $T$ feature planes represent the occupancy of the player's discs, with a feature value of 1 indicating that a disc occupies the corresponding cell, and 0 otherwise. Similarly, the following $T$ feature planes represent the occupancy of the opponent's discs. The final plane encodes the current player, being filled with $+1$ when it is the first player's turn and with $-1$ otherwise.

\subsection{Monte Carlo Tree Search in AlphaZero}

This subsection provides an explanation of the Monte Carlo Tree Search (MCTS) algorithm used in AlphaZero. Each node in the game tree represents a game state, and each edge $(s, a)$ represents a valid action from that state. The edges store a set of statistics: $\{N(s, a), W(s, a), Q(s, a), p(s, a)\}$, where $N(s, a)$ is the visit count, $W(s, a)$ is the cumulative value, $Q(s, a) = W(s, a)/N(s, a)$ is the mean value, and $p(s, a)$ is the move probability.

The MCTS for AlphaZero consists of four steps: {\it Select}, {\it Expand and Evaluate}, {\it Backup}, and {\it Play}. A simulation is defined as a sequence of {\it Select}, {\it Expand and Evaluate}, and {\it Backup} steps, repeated $N_\mathrm{sim}$ times. {\it Play} is executed after $N_\mathrm{sim}$ simulations.

In {\it Select}, the tree is searched from the root node $s_\mathrm{root}$ to the leaf node $s_L$ at time step $L$ using a variant of the PUCT algorithm. At each time step $t < L$, the selected action $a_t$ has the maximum score, as described in Eq. \ref{eq:uct}:
\begin{equation}
  \label{eq:uct}
 a_t = \argmax_a (Q(s_t, a) + C_\mathrm{puct} p(s_t, a) \frac{\sqrt{N(s_t)}}{1 + N(s_t, a)}),
\end{equation}
where $N(s_t)$ is the number of parent visits and $C_\mathrm{puct}$ is the exploration rate. In this study, $C_\mathrm{puct}$ is constant, whereas in the original AlphaZero, $C_\mathrm{puct}$ increases slowly with search time \cite{Silver:2018}. Additionally, when the parent node is the root node, the node selection is performed using an $\epsilon$-greedy algorithm based on the UCT scores.

In {\it Expand and Evaluate}, the DNN evaluates the leaf node and outputs $v(s_L)$ and $\bm p(s_L)$.
If the leaf node is a terminal node, $v(s_L)$ is the color of the winning player's disc.
The leaf node is expanded and each edge $(s_L, a)$ is initialized to $\{N(s_L, a) = 0, W(s_L, a) = 0, Q(s_L, a) = 0, p(s_L, a) = p(a \mid s_L)\}$.

In {\it Backup}, the visit counts and values are updated for each step $t \leq L$ during the backward pass.
The visit count is incremented by 1, $N(s_t, a_t) \leftarrow N(s_t, a_t) + 1$, and the cumulative and average values are updated, $W(s_t, a_t) \leftarrow W(s_t, a_t) + v$, $Q(s_t, a_t) \leftarrow W(s_t, a_t) /N(s_t, a_t)$.

Finally, in {\it Play}, AlphaZero selects the action corresponding to the most visited edge from the root node.

\section{Training}
\label{sec:training}

AlphaViT, AlphaViD, AlphaVDA, and AlphaZero share the same three-stage training loop: \textit{Self-play}, \textit{Augmentation}, and \textit{Update}. One complete cycle of the three stages is called an \textbf{iteration} and is repeated $N_{\mathrm{iter}}$ times. This training algorithm is a modified version of the original AlphaZero, adapted for a single-machine setting.

During the {\it Self-play} phase, an agent plays against itself $N_\mathrm{self}$ times. For the first $T_\mathrm{opening}$ turns, actions are stochastically selected among the valid moves according to the softmax policy defined in Eq. \ref{eq:softmax}:
\begin{equation}
  \label{eq:softmax}
 p(a \mid s) = \exp(N(s,a)/\tau)/\sum_b \exp(N(s,b)/\tau),
\end{equation}
where $\tau$ is a temperature parameter that controls the exploration. This stochastic exploration enables the agent to explore new and potentially better actions. After $T_\mathrm{opening}$, the most visited action is selected. During \textit{Self-play}, we record the board states, game outcomes (winners), and the search probabilities. The search probabilities represent the probabilities of selecting valid moves at the root node in MCTS.

In the {\it Augmentation} phase, the dataset derived from {\it Self-play} is augmented by introducing symmetries specific to the game variant (e.g., two symmetries for Connect 4 and eight for Othello and Gomoku). This augmented data is added to a queue with a capacity of $N_\mathrm{queue}$ states to form the training dataset.

For the first update iteration, the training data queue is filled with data generated by self-play using MCTS100, which is then augmented. In subsequent iterations, new data generated by {\it Self-play} are added to the training data queue. To simultaneously learn multiple games, we prepare a separate training data queue for each game.

During the {\it Update} phase, the DNN is trained using mini-batch stochastic gradient descent with a batch size of $N_{\mathrm{batch}}$ for $N_{\mathrm{epochs}}$ epochs. The optimizer uses weight decay. The loss function $l$ combines the mean squared error between the predicted value $v$ and the winner's disc color $c_\mathrm{win}$, and the cross-entropy loss between the search probabilities $\bm{\pi}$ and the predicted move probabilities $\bm{p}$. The loss function is defined in Eq. \ref{eq:loss}:
\begin{equation}
  \label{eq:loss}
 l = (c_\mathrm{win} - v)^2 - \bm \pi^\mathrm{T} \log \bm p.
\end{equation}

To train multiple games simultaneously, mini-batches are generated from the respective training data queue of each game. During the {\it Update} phase, mini-batches are sampled from these individual queues and used to update the DNN. For example, when an agent simultaneously trains Connect 4, Gomoku, and Othello, we sequentially use one mini-batch from the training data queues of Connect 4, Gomoku, and Othello to update the network.

\section{Parameters}
\label{sec:parameters}

The hyperparameters for AlphaViT, AlphaViD, AlphaVDA, and AlphaZero are listed in Table \ref{tab:param_alphavit}. For training, we employ the AdamW optimizer in PyTorch, with all parameters set to their default values except for the learning rate. All other parameters for AlphaZero were consistent with the previous implementation \citep{Fujita:2022}. The hyperparameters of the other models were carefully hand-tuned to optimize their performance. 

Table \ref{tab:param_alphavit_game} lists the game-specific hyperparameters for AlphaViT, AlphaViD, AlphaVDA, and AlphaZero. The number of MCTS simulations ($N_\mathrm{sim}$) ranges from 200 to 400, depending on the game and board size. The number of self-play games per iteration is set to 30 for Connect 4 variants and 10 for Gomoku and Othello variants. The opening phase ($T_\mathrm{opening}$) specifies the number of initial moves using softmax decision-making with a temperature parameter ($\tau$) that is adjusted based on the game and board size.

\begin{table}[htbp]
  \centering
  \caption{Hyperparameters of AlphaViT, AlphaViD, AlphaVDA, and AlphaZero}
  \label{tab:param_alphavit}
  \begin{tabular}{|c|c|c|c|c|} \hline
    Parameter                     &      AlphaViT &     AlphaViD &       AlphaVDA & AlphaZero \\ \hline
    $C_\mathrm{puct}$             &          1.25 &         1.25 &          1.25 &     1.25  \\ \hline
    $\varepsilon$                 &           0.2 &          0.2 &           0.2 &      0.2  \\ \hline 
    $T$                           &             1 &            1 &             1 &        1  \\ \hline 
    $N_\mathrm{queue}$            &        100000 &       100000 &        100000 &   100000  \\ \hline 
    $N_\mathrm{epoch}$            &             1 &            1 &             1 &        1  \\ \hline
    Optimizer                     &         AdamW &        AdamW &         AdamW &     AdamW  \\ \hline
    Batch size                    &          1024 &         1024 &          1024 &     1024  \\ \hline    
    Learning rate                 &        0.0001 &       0.0001 &        0.0001 &   0.0001   \\ \hline
    Weight decay                  &          0.01 &         0.01 &          0.01 &     0.01  \\ \hline
    Patch size                    &   $5\times5$  &    $5\times5$ &    $5\times5$ &        -  \\ \hline 
    Patch stride                  &            1  &            1 &             1 &        -  \\ \hline 
    Embedding size of encoder     &           512 &          512 &           512 &        -  \\ \hline
    Encoder feedforward dimension &          1024 &         1024 &          1024 &        -  \\ \hline
    Number of encoder heads       &             8 &            8 &             8 &        -  \\ \hline
    Size of positional embeddings & $512\times256$& $512\times256$& $512\times256$&        -  \\ \hline
    Number of decoder layers      &            -  &            1 &             1 &        -  \\ \hline
    Embedding size of decoder     &            -  &          512 &           512 &        -  \\ \hline
    Decoder feedforward dimension &            -  &         1024 &          1024 &        -  \\ \hline
    Number of decoder heads       &            -  &            8 &             8 &        -  \\ \hline
    Size of action embeddings     &            -  &            - &           256 &        -  \\ \hline
    Dropout rate                  &           0.1 &          0.1 &           0.1 &        -  \\ \hline
    Number of residual blocks     &            -  &            - &             - &        6  \\ \hline
    Kernel size                   &            -  &            - &             - &        3  \\ \hline
    Number of filters             &            -  &            - &             - &      256  \\ \hline
  \end{tabular}
\end{table}

\begin{table}[htbp]
  \centering
  \caption{Game-specific hyperparameters for AlphaViT, AlphaViD, AlphaVDA, and AlphaZero}
  \label{tab:param_alphavit_game}
  \begin{tabular}{|c|c|c|c|c|c|c|} \hline
                          & Connect 4 & Connect 4 5x4&  Gomoku &Gomoku 6x6 & Othello &  Othello 6x6 \\ \hline
    Number of simulations &      200 &         200 &     400 &      200 &     400 &        200  \\ \hline
    Number of self-play   &       30 &          30 &      10 &       10 &      10 &         10  \\ \hline
    $T_\mathrm{opening}$  &        4 &           4 &       8 &        6 &       8 &          6  \\ \hline 
    $\tau$                &      100 &         100 &      40 &       20 &      80 &         40  \\ \hline 
  \end{tabular}
\end{table}

\section{Monte Carlo tree search}
\label{sec:mcts}

MCTS is a sampling-based tree-search algorithm that does not require a predefined evaluation function \citep{Browne:2012,Winands:2017}. It operates through four strategic steps: {\it Selection step}, {\it Expansion step}, {\it Playout step}, and {\it Backpropagation step} \citep{Winands:2017}. In the {\it Selection step}, the tree is traversed from the root node to a leaf node. Child node $c$ of parent node $p$ is selected to maximize the score defined in Eq. \ref{eq:uct_mcts}:
\begin{equation}
  \label{eq:uct_mcts}
 \mathrm{UCT} = q_c / N_c + C \sqrt{\frac{2\ln(N_p + 1)}{N_c + \varepsilon}},
\end{equation}
where $q_c$ is the cumulative value of $c$, $N_p$ is the visit count of $p$, $N_c$ is the visit count of $c$, and $\varepsilon$ is a constant value to avoid division by zero. In the {\it Expansion step}, child nodes are added to the leaf node when the visit count of the leaf node reaches $N_\mathrm{open}$. In the {\it Playout step}, a valid move is selected at random until the end of the game is reached. In the {\it Backpropagation step}, the result of the playout is propagated along the path from the leaf node to the root node. If the MCTS player itself wins, loses, and draws, $q_i \leftarrow q_i + 1$, $q_i \leftarrow q_i - 1$, and $q_i \leftarrow q_i$, respectively. $q_i$ is the cumulative value of node $i$. The sequence of these four steps constitutes a single simulation. The simulation process is repeated $N_\mathrm{sim}$ times. After all simulations, MCTS selects the action corresponding to the most visited child node of the root. In this study, $C = 0.5$, $\varepsilon = 10^{-7}$, and $N_\mathrm{open} = 5$.

\section{Minimax algorithm}
\label{sec:minimax}

The minimax algorithm is a fundamental game-tree search technique that determines the optimal action by evaluating the best possible outcome for the current player. Each node in the tree contains a state, player, action, and value. The algorithm creates a game tree with a depth of $N_\mathrm{depth}=3$. For the Othello variants, the algorithm expands the tree to the terminal nodes after the last six turns. The root node corresponds to the current state and minimax player. Next, the states corresponding to leaf nodes are evaluated. Then, the algorithm propagates the values from the leaf nodes to the root node. If the player corresponding to the node is the opponent, the value of the node is the minimum value of its child nodes. Otherwise, the value of the node is the maximum value of its child nodes. Finally, the algorithm selects the action corresponding to the root's child node with the maximum value. The evaluation of leaf nodes is tailored to each game.

For the Connect 4 variants, the values of the connections of two and three same-colored discs are $R\times c_\mathrm{disc} c_\mathrm{minimax}$ and $R^2\times c_\mathrm{disc} c_\mathrm{minimax}$, respectively, where $R$ is the base reward, and $c_\mathrm{disc}$ and $c_\mathrm{minimax}$ are the colors of the connecting discs and the minimax player's disc, respectively. The value of a node is the sum of the values of all connections on the corresponding board. The terminal nodes have a value of $R^3 c_\mathrm{win} c_\mathrm{minimax}$, where $c_\mathrm{win}$ is the color of the winner, and $R=100$.

For the Gomoku variants, the values of the connections of two, three, and four same-colored discs are $R\times c_\mathrm{disc} c_\mathrm{minimax}$, $R^2\times c_\mathrm{disc} c_\mathrm{minimax}$, and $R^3\times c_\mathrm{disc} c_\mathrm{minimax}$, respectively. The value of a node is the sum of the values of all connections on the corresponding board. The terminal nodes have values of $R^4 c_\mathrm{win} c_\mathrm{minimax}$ and $R=100$.

For the Othello variants, the value of a node is calculated using Eq. \ref{eq:evaluation_minimax}.
\begin{equation}
  \label{eq:evaluation_minimax}
 E = \sum_x \sum_y v(x,y) o(x,y) c_\mathrm{minimax},
\end{equation}
where $v(x, y)$ is the value of cell $(x, y)$ and $o(x, y)$ is the occupancy of cell $(x, y)$. For Othello 6x6 and Othello, the minimax algorithm evaluates each cell using Eq. \ref{eq:evaluation_othello66} and \ref{eq:evaluation_othello88}, respectively. $o(x, y)$ is $1$, $-1$, and $0$ if cell $(x, y)$ is occupied by the first player's disc, the second player's disc, and empty, respectively. The terminal nodes have a value of $E_\mathrm{end} = 1000 c_\mathrm{win} c_\mathrm{minimax}$.

\begin{equation}
    \label{eq:evaluation_othello66}
    v_{6\times6} = \left(
    \begin{matrix}
        30&  -5&   2&   2&  -5& 30\\
        -5& -15&   3&   3& -15& -5\\
         2&   3&   0&   0&   3&  2\\
         2&   3&   0&   0&   3&  2\\
        -5& -15&   3&   3& -15& -5\\
        30&  -5&   2&   2&  -5& 30\\
\end{matrix}
\right).
\end{equation}
\begin{equation}
    \label{eq:evaluation_othello88}
    v_{8\times8} = \left(
    \begin{matrix}
        120& -20&  20&   5&   5&  20& -20& 120\\
        -20& -40&  -5&  -5&  -5&  -5& -40& -20\\
         20&  -5&  15&   3&   3&  15&  -5&  20\\
          5&  -5&   3&   3&   3&   3&  -5&   5\\
          5&  -5&   3&   3&   3&   3&  -5&   5\\
         20&  -5&  15&   3&   3&  15&  -5&  20\\
        -20& -40&  -5&  -5&  -5&  -5& -40& -20\\
        120& -20&  20&   5&   5&  20& -20& 120\\
\end{matrix}
\right).
\end{equation}

\section{Elo rating}
\label{sec:elo}

\subsection{Calculation of Elo rating}

Elo rating is a widely used metric for evaluating the relative performance of players in two-player games. This metric allows us to estimate the probability of one player defeating another based on their current ratings. Given two players A and B with Elo ratings $e(A)$ and $e(B)$, respectively, the probability that player A will defeat player B, denoted $p(A \mathrm{ defeats } B)$, is calculated using Eq. \ref{eq:elo}:
\begin{equation}
\label{eq:elo}
 p(A ~\mathrm{defeats}~ B) = 1/(1 + 10^{(e(B) - e(A))/400} ).
\end{equation}
After a series of $N_G$ games between players A and B, player A's Elo rating is updated to a new value $e'(A)$ based on their performance. The update is performed using Eq. \ref{eq:elo_update}:
\begin{equation}
  \label{eq:elo_update}
 e'(A) = e(A) + K(N_\mathrm{win} - N_G \times p(A ~\mathrm{defeats}~ B)),
\end{equation}
where $N_\mathrm{win}$ denotes the total score accumulated by player A across the $N_G$ games (counting each win as 1, each draw as 0.5, and each loss as 0), and $K$ is a factor that determines the maximum rating adjustment after a single game. In this study, $K = 8$.

\subsection{Calculation of 95\% confidence intervals for Elo ratings}

To quantify estimation uncertainty, we compute 95\% confidence intervals via percentile bootstrap \citep{LMSYS:2023,Tang:2025}:
\begin{enumerate}[label=(\arabic*)]
  \item Resample the full set of $T$ game results with replacement from the original dataset (which contains $T$ samples), yielding a new dataset of size $T$.
  \item Apply the same sequential-update Elo rule (initial rating = 1500, $K = 8$) to compute ratings for all players.
  \item Repeat steps 1 and 2 for $B = 1000$ replicates to obtain a distribution of Elo ratings.
  \item Define the 95\% confidence interval for each player as the 2.5th and 97.5th percentiles of its bootstrap Elo distribution.
\end{enumerate}
This yields point estimates from the original data and interval estimates from the bootstrap replicates, enhancing the statistical rigor of the evaluation.

\subsection{Calculation of 95\% confidence intervals for ratio of Elo ratings between deeper and shallower models}

We compute an Elo rating for each training iteration ($t$) by replaying the complete match log in chronological order. For player $p$, we obtain an iteration series ${E_{p,t}} (t =$ 1,2, $\ldots$,10,20,40, $\ldots$,100,200,300, $\ldots$,3000$)$. To summarize playing strength over the final one thousand iterations, we consider the ten checkpoints in the interval $T=\{2100, 2200, \dots, 3000\}$. The mean Elo in this window is $\bar{E}_{p,s}=\frac{1}{|T|}\sum_{t \in T}E_{p,t}$, where ($s\in\{\text{shallower},\text{deeper}\}$) labels the shallower and deeper variants of the same architecture. From these averages we form the relative performance ratio $R=\bar{E}_{p,\text{deep}} / \bar{E}_{p,\text{shallow}}$. Additionally, we also compute the difference in average Elo ratings $\Delta=\bar{E}_{p,\text{deep}}-\bar{E}_{p,\text{shallow}}$. Sampling uncertainty for $R$ is quantified using a non-parametric percentile bootstrap: the ten checkpoint rows are resampled \(B=1000\) times with replacement, the bootstrap statistics ($\{R^{*b}\}_{b=1}^{B}$) are recomputed, and a 95\% confidence interval for $R$ is derived from the 2.5th and 97.5th percentiles of the resulting empirical distribution.

\section{Additional AlphaZero experiments}
\label{sec:adamw}

These experiments systematically evaluated how architectural depth and optimization methods influenced the learning dynamics and performance of AlphaZero models. The baseline configuration for this study, denoted as Res6, utilized a neural network architecture comprising six residual blocks trained with the AdamW optimizer. The deeper variant, Res10, increased the architectural depth to ten residual blocks and similarly employed AdamW optimization. A further variant, Res10 fine-tuning, represented the Res10 model refined through fine-tuning by employing the same procedure described in Subsection \ref{sec:fine-tuning}. Notably, before fine-tuning, only the heads were initialized with random weights because these heads for the small board configuration were not utilized and were replaced with new heads suitable for large board configurations. Finally, to specifically investigate the impact of optimizer selection, the Res6 SGD configuration maintained the six-block structure of Res6, but substituted AdamW with stochastic gradient descent (SGD). The SGD hyperparameters (the learning rate, weight decay, and momentum) were set to 0.05, 0.0001, and 0.9, respectively. The training process for Res6 SGD was identical to that denoted in Appendix \ref{sec:training}, except for the optimizer choice. Elo ratings were calculated according to the methodology detailed in Subsection \ref{sec:variation_elo}.

\begin{figure}
  \begin{center}
      \includegraphics[width=1\linewidth]{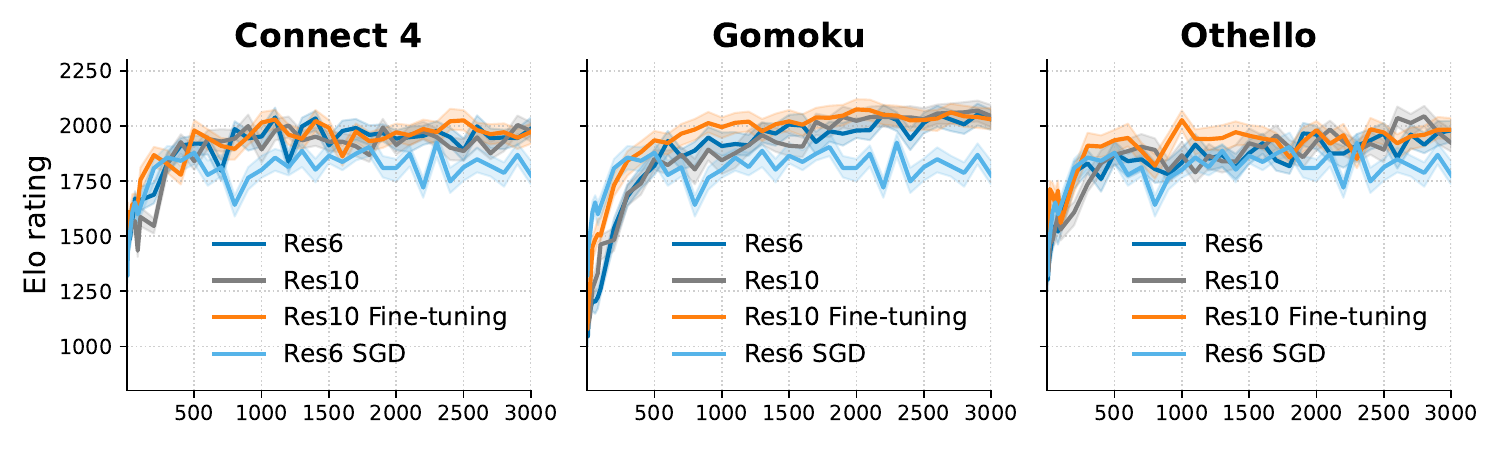}
      \caption{Elo rating progression over training iterations for AlphaZero variants on large board configurations in Connect 4, Gomoku, and Othello. Res6: AlphaZero with six residual blocks. Res10: AlphaZero with ten residual blocks. Res10 fine-tuning: Res10 with fine-tuning. Res6 SGD: Res6 trained using the SGD optimizer.}
      \label{fig:elo_alphazero}                
  \end{center}
\end{figure}

Figure \ref{fig:elo_alphazero} illustrates the progression of Elo ratings over training iterations for the AlphaZero configurations across three games: Connect 4, Gomoku, and Othello, in large board scenarios. The results indicate that all AlphaZero variants achieve high Elo ratings with sufficient training. Notably, the Res10 fine-tuning configuration exhibits a rapid initial improvement in Gomoku, quickly attaining a high Elo rating relative to the non-fine-tuned Res10 configuration. Conversely, the Res6 SGD configuration consistently demonstrates inferior performance compared to its AdamW-optimized counterpart, plateauing prematurely and failing to achieve comparable Elo ratings across all games.

Collectively, these findings underscore the efficacy of Res6 architecture optimized with AdamW, affirming its suitability as a stable baseline for comparative studies. Moreover, AlphaZero demonstrates notable efficiency in terms of parameter count and computational resource requirements relative to transformer-based models, such as AlphaViT, AlphaViD, and AlphaVDA. In addition, deeper AlphaZero architectures benefit significantly from fine-tuning, rapidly achieving strong performance with reduced computational overhead.

\section{Additional MCTS experiment}
\label{sec:mcts_experiment}

We conducted an additional experiment to investigate the impact of varying simulation counts on the performance of MCTS in Connect 4, using the large board configuration. The MCTS algorithm was systematically evaluated using a fixed number of simulations ranging from 10 to 6400. Each MCTS variant was evaluated against the set of agents previously described (see Table \ref{tab:Connect4_elo}), with the opponents' Elo ratings maintained at the values already established. Elo ratings for the MCTS variants were calculated using the method described in Subsection \ref{sec:variation_elo}.

The results of these evaluations are presented in Table~\ref{tab:mcts}, indicating a significant improvement in Elo ratings as the simulation count increases from 10 to 1600. Beyond 1600, the incremental Elo gain becomes less pronounced; 6400 adds only $\approx 30$ Elo, and 3200 shows no improvement, illustrating diminishing returns. This observed trend aligns closely with findings reported by \cite{Baier:2015}, who demonstrated that enhancing the inference time, conceptualized as the number of simulations in MCTS, does not necessarily lead to continuous performance improvements in Connect 4. Furthermore, we observe a substantial difference in performance between 100- and 400-simulation configurations, which are selected as baseline configurations. 

\begin{table}[htbp]
  \centering
  \caption{Elo ratings of MCTS variants in Connect 4}
  \label{tab:mcts}
  \begin{tabular}{|l|c|c|}
  \hline
  Simulations & Elo rating & 95\% CI \\
  \hline
    10  & 1048.3   & [1004.0, 1093.1] \\
   100  & 1316.8   & [1273.3, 1366.4] \\
   200  & 1431.4   & [1376.5, 1472.3] \\
   400  & 1509.8   & [1476.3, 1564.8] \\
   800  & 1573.6   & [1522.8, 1623.3] \\
  1600  & 1676.7   & [1611.0, 1715.0] \\
  3200  & 1673.5   & [1624.3, 1730.7] \\
  6400  & 1708.4   & [1674.1, 1767.1] \\
  \hline
  \end{tabular}
\end{table}

\section{Computational performance}
\label{sec:performance}

Tables \ref{tab:computational_performance_training} and \ref{tab:computational_performance_inference} summarize the computational performance of each model for Othello, including the number of parameters, peak GPU memory consumption, and mean execution time during training and inference, respectively. All experiments were conducted on a custom-made computer equipped with an Intel Core i9-12900K CPU, 64 GiB of DDR5 RAM, and two NVIDIA RTX 4060 Ti 16 GiB GPUs. For training, we employed a data-parallel approach utilizing both GPUs and ten parallel self-play games (five per GPU). During inference, a single process on a single GPU was used. To measure the training and inference speeds, two precision modes were considered: automatic mixed precision (FP16) and full precision (FP32). The experimental setup included PyTorch 2.3.1, NVIDIA Driver Version 560.35.03, and CUDA Version 12.6.

Peak GPU memory usage (in MiB) denotes the highest memory allocation recorded during training and inference. During training, peak memory usage represents the combined memory usage across both GPUs. Training speed was reported as seconds per iteration (s/iteration), where an iteration comprised one complete cycle of self-play, data augmentation, and network updates (as detailed in Appendix \ref{sec:training}). Inference speed was measured in seconds per decision (s/decision), reflecting the time required to compute one move. Mean values for training and inference speeds were computed over ten training iterations and ten full games played by the agent against itself, respectively. 

The results demonstrate that for training, FP16 generally provides faster performance and reduced memory usage compared to FP32 for AlphaViT, AlphaViD, and AlphaVDA, but not for AlphaZero. Conversely, during inference, FP32 is generally faster and more memory-efficient than FP16 for all models. Consequently, FP16 was used for training and FP32 was used for inference in this study.

Additionally, we employed the number of model parameters and other hyperparameters (e.g., number of games per iteration and batch size) listed in Table \ref{tab:param_alphavit_game} for training and evaluating the agents because our computational resources were limited to a single computer with two GPUs, and we were able to train and evaluate the agents with practical computation time for this research. 

It is important to note that computation time is significantly influenced not only by model size but also by the game engine implementation, hardware specifications, and hyperparameters, such as the number of parallel processes for self-play and the number of simulations. Achieving a practical balance between these factors is crucial to ensure efficient training and inference.

Initially, for each proposed architecture, we selected the shallowest encoder depth (4 encoder layers for AlphaViT and one encoder layer for AlphaViD/AlphaVDA) that resulted in approximately 11 million parameters. This depth was chosen to achieve a balanced trade-off between model size, computational efficiency, and expected performance. However, for large board configurations, initial results indicated that these shallower models (especially AlphaViD and AlphaVDA) did not exhibit competitive performance against the AlphaZero baseline. Consequently, we introduced deeper models—AlphaViT with eight encoder layers (AlphaViT L8) and AlphaViD/AlphaVDA with five encoder layers (AlphaViD L5, AlphaVDA L5)—resulting in approximately 20 million parameters. These deeper configurations represented practical upper limits given our available computational resources, in particular the memory and performance constraints of our hardware setup consisting of two NVIDIA RTX 4060 Ti GPUs, each with 16 GiB of GPU memory.

\begin{table}[htbp]
  \centering
  \caption{Computational performance during training}
  \begin{tabular}{|l|c|c|c|c|c|c|c|}
    \hline
    Model                   & Params (M) & \multicolumn{2}{c|}{Peak Mem (MiB)} & \multicolumn{2}{c|}{Mean time (sec/iteration)}   \\
                            &            & FP16  & FP32  & FP16   & FP32 \\
    \hline
    AlphaViT L4             & 11.2       &  6672 &  9582 & 131.2 & 132.5 \\
    AlphaViT L8             & 19.6       & 11260 & 17292 & 197.3 & 208.8 \\
    AlphaViD L1             & 11.5       &  4980 &  7866 & 113.7 & 114.1 \\
    AlphaViD L5             & 19.9       &  9390 & 14902 & 180.6 & 189.1 \\
    AlphaVDA L1             & 11.3       &  4540 &  7074 & 109.6 & 110.6 \\
    AlphaVDA L5             & 19.8       &  9232 & 14598 & 177.8 & 184.9 \\
    AlphaZero (6 ResBlocks) & 7.1        &  3526 &  3107 & 112.2 & 102.9 \\
    AlphaZero (10 ResBlocks)& 11.8       &  5262 &  4448 & 149.4 & 134.3 \\
    \hline
  \end{tabular}
  \label{tab:computational_performance_training}
  \caption*{\
    Peak memory is the maximum aggregated GPU usage observed over ten iterations. Mean iteration time (second/iteration) is averaged over the same interval.
  }
\end{table}

\begin{table}[htbp]
  \centering
  \caption{Computational performance during inference}
  \begin{tabular}{|l|c|c|c|c|c|c|c|}
    \hline
    Model                   & Params (M) & \multicolumn{2}{c|}{Peak Mem (MiB)} & \multicolumn{2}{c|}{Mean time (sec/decision)}   \\
                            &            & FP16  & FP32  & FP16   & FP32 \\
    \hline
    AlphaViT L4             & 11.2       & 253   & 237   & 0.7541 & 0.6546 \\
    AlphaViT L8             & 19.6       & 307   & 261   & 1.038  & 0.6763 \\
    AlphaViD L1             & 11.5       & 251   & 239   & 0.7104 & 0.5802   \\
    AlphaViD L5             & 19.9       & 307   & 263   & 1.006  & 0.8018   \\
    AlphaVDA L1             & 11.3       & 249   & 237   & 0.7116 & 0.5943   \\
    AlphaVDA L5             & 19.8       & 305   & 261   & 0.9689 & 0.8199   \\
    AlphaZero (6 ResBlocks) & 7.1        & 233   & 207   & 0.7864 & 0.7245   \\
    AlphaZero (10 ResBlocks)& 11.8       & 273   & 227   & 0.9404 & 0.8293   \\
    \hline
  \end{tabular}
  \label{tab:computational_performance_inference}
  \caption*{
    Peak memory is the maximum single-GPU usage over ten self-play games. Mean decision time is averaged over the same games.
  }
\end{table}

\clearpage


\end{document}